\documentclass[11pt]{article}
\usepackage{geometry}
\usepackage{array}
\usepackage{float}
\usepackage{multicol}
\usepackage{multirow}
\usepackage[affil-it]{authblk}
\usepackage{hyperref}
\usepackage{appendix}
\usepackage{verbatim}
\usepackage{epstopdf}
\usepackage{graphicx}
\usepackage{amssymb}
\usepackage{amsmath}
\usepackage{bm}
\usepackage{subcaption}
\usepackage{rotating}
\usepackage{lipsum}
\usepackage{caption}
\usepackage{makeidx}
\usepackage[english]{babel}
\usepackage{cite}
\usepackage{authblk}	

\newcommand{\etal}{\textit{et al}.}
\newcommand{\ie}{\textit{i}.\textit{e}.}
\newcommand{\eg}{\textit{e}.\textit{g}.}
\newcommand{\Vol}{\rotatebox[origin=c]{180}{\ensuremath{A}}}

\makeindex

\begin{document}

\title{\bf Hull Form Optimization with Principal Component Analysis and Deep Neural Network}
   
\author{Dongchi Yu\thanks{birdmanyu@berkeley.edu} }
\author{Lu Wang\thanks{lu$\_$wang$\_$0405@berkeley.edu}}
\affil{Department of Mechanical Engineering, \\ University of California at Berkeley}
   
\date{}
\maketitle

\begin{abstract}
Designing and modifying complex hull forms for optimal vessel performances have been a major challenge for naval architects. In the present study, Principal Component Analysis (PCA) is introduced to compress the geometric representation of a group of existing vessels, and the resulting principal scores are manipulated to generate a large number of derived hull forms, which are evaluated computationally for their calm-water performances. The results are subsequently used to train a Deep Neural Network (DNN) to accurately establish the relation between different hull forms and their associated performances. Then, based on the fast, parallel DNN-based hull-form evaluation, the large-scale search for optimal hull forms is performed. 
\end{abstract}

\section{Introduction}
Obtaining optimal designs of ship hulls that meet technical and operational requirements has always been a major challenge for naval architects. In particular, the complex geometric shapes of vessels and the non-trivial relation between hull shape and vessel performance lead to a time-consuming iterative design process. 

Direct experimental and computational hydrodynamic analyses have long been performed to evaluate the performances of ships of various sizes and shapes. Important aspects of vessel performance such as calm-water resistance~\cite{wehausen1973, dawson1977, raven1996, yeung2004, yu2017omae}, seakeeping~\cite{korvin1957, yeung1982, faltinsen1993}, and maneuverability~\cite{yeung1978, inoue1981, bertram2012} have been widely addressed by numerous hydrodynamic studies. Various theoretical, computational, and experimental modeling approaches are adopted in the process, such as potential-flow modeling~\cite{wehausen1960, newman1977}, viscous-fluid Computational Fluid Dynamic (CFD) simulations~\cite{ferziger2012}, and model-scale experiments~\cite{souto2007, kim2014, yu2017}. In these investigations, the complex hull surfaces are represented either by computational meshes or physical models, which are generally difficult to generate and manipulate. Therefore, although direct hydrodynamic analysis provides a detailed description of the flow behavior around the hull and the vessel performance, the associated high cost of computation and experimentation makes it impractical to evaluate a large amount of different hull forms for optimization purposes.

In practice, naval architects frequently seek design guidance from existing hull forms. Relevant approaches include line distortion~\cite{bertram1998, han2012} and standard series analysis~\cite{molland2011, papanikolaou2014}, in which the target hull form is acquired by applying slight adjustments to the lines plans of one or multiple reference hulls. In other studies, the surface of a vessel is represented by parametric splines, and optimization algorithms such as the Evolution Algorithm~\cite{dejhalla2001, grigoropoulos2010} are introduced to iteratively adjust the parametric representation of the vessel for the best hydrodynamic performance based on CFD computational results. However, this design approach only allows moderate changes being made to the parent hulls, and CFD computation is invoked at each step of the design iteration, which makes the optimization process very time-consuming. Furthermore, the entire design process must be performed again when design requirements change.

Alternatively, simple statistic analyses may also be utilized to establish the relation between hydrodynamic performance of interest and certain key geometric coefficients based on knowledge of existing vessels~\cite{fairlie1975, holtrop1982, watson2002}, so that the general sizes and shapes of the hulls can be quickly determined at an early stage. However, it is hard to accurately describe the complicated surface geometries of vessels using only a handful of selected geometric coefficients, and the highly nonlinear relation between vessel performance and hull geometry is difficult to accurately approximate by simple statistical approaches such as linear regression.

The advent of the ``big data era"~\cite{john2014} has provided newer and more sophisticated data-assisted approaches for hull design and optimization. In particular, with the modern deep-learning programming framework~\cite{tensorflow2015-whitepaper, abadi2016} and high-performance computation, machine-learning techniques can be easily applied to identify and approximate complicated patterns of very large datasets with relatively low computing costs. Existing applications of machine learning on naval architecture and ocean engineering can be found on the various topics such as ocean monitoring~\cite{tsai2002, jain2006}, marine structure optimization~\cite{cui2012}, and engine control~\cite{perera2016}. Still, although machine learning has been successfully implemented in many areas~\cite{nasrabadi2007}, its full potential is yet to be tapped into by the ship design community. In particular, the application of machine learning on the design and optimization of hull forms is rarely, if ever, explored.

Therefore, in the present study, we propose a novel design approach based on a combination of hydrodynamic analysis and machine learning techniques to improve hull-form geometry for the optimal calm-water transporting efficiency. Principal Component Analysis (PCA) is performed first to acquire the reduced-dimensionality representations of a group of parent hulls, so that the complicated hull forms can be fully represented by and recovered from several principal scores. Then, these principal scores are manipulated to generate a large number of hull geometries, and their associated calm-water efficiencies are evaluated by an efficient potential-flow boundary-element method. The computational results are then used to train and test a Deep Neural Network (DNN) to accurately approximate the relation between hull forms, forward speeds, and the associated calm-water performance. Finally, based on the fast, parallel forward propagation on the trained DNN, a large amount of hull forms can be quickly evaluated, so that the optimal hull form with the best calm-water efficiency can be identified. Since no additional hydrodynamic computation is required in the DNN-based hull-form evaluation, large-scale search for the optimal hull form can be performed at a much lower cost with similar accuracy as that of the training samples. In addition, design requirements such as operating speed and general dimensions of the hull can be easily incorporated into the DNN-based search. The final optimal design can, therefore, be generated by suitably integrating the various geometric features of the parent hulls for the target design and operational requirements.

\section{Principal Component Analysis (PCA) with Application to Hull-Form Representation} 
\label{sec:2}
Principal Component Analysis (PCA) is a popular dimensionality reduction algorithm which attempts to capture most of the variance in a high-dimensional dataset with a low-dimensional representation~\cite{shalev2014}. PCA compresses an otherwise high-dimensional data point into a handful of principal scores, while the original data point can still be recovered with limited error from the principal scores using the transformation matrix. Such a feature can be very useful in handling complex vessel geometries, and the procedure of the proposed approach is briefly introduced in this section.

\subsection{Brief Introduction of PCA}
For a given dataset in which each sample is zero centered, let $\mathbf{x}_1,\dots,\mathbf{x}_n$ be $n$ column vectors in $\mathbb{R}^l$ representing $n$ samples with $l$ features, and constitute a data matrix $\mathbf{X} \in \mathbb{R}^{n, l}$, \ie
\begin{equation}
\label{eq:2.1}
\mathbf{X} = [\mathbf{x}_1, \mathbf{x}_2, \ldots , \mathbf{x}_n]^\intercal.
\end{equation}

\noindent Then, a transformation matrix $\mathbf{W}\in\mathbb{R}^{l,l}$ is introduced to map each sample vector $\mathbf{x}_i$ into a vector of $l$ principal scores $\mathbf{t}_i\in\mathbb{R}^l$, and the full transformation of $\mathbf{X}$ into a new matrix of principal scores $\mathbf{T}\in\mathbb{R}^{n,l}$ is given by

\begin{equation}
\label{eq:2.2}
\mathbf{T} = [\bm{t}_1,\ldots,\bm{t}_n]^\intercal = \mathbf{X}\mathbf{W}.
\end{equation}

\noindent Consequently, a new $l$-dimensional space is constructed, and each of the dimensions is arranged in such a way to successively inherit the maximum possible variance from the original data matrix $\mathbf{X}$, \ie~the first value of $\mathbf{t}_i$, $t_{i,1}$ accounts for the largest variance for the given dataset $\mathbf{X}$, and $t_{i,2}$ accounts for the second largest variance and so on. Matrix $\mathbf{W}$ can be acquired through various means such as Singular Value Decomposition (SVD)~\cite{golub1970}, which is discussed in details in the literatures~\cite{wall2003, abdi2010}. 

For datasets of practical interest, it is frequently discovered that the first few columns of matrix $\mathbf{T}$ account for a predominant portion of the variance of the original dataset; therefore, the dimensionality of the original dataset can be greatly reduced by discarding the columns of $\mathbf{T}$ with little variance, while most of the information of the original data matrix is still preserved. For instance, the first $d$ principal components of $\mathbf{X}$ can be extracted with

\begin{equation}
\label{eq:2.3}
\mathbf{\tilde{T}} = \mathbf{X}\mathbf{\tilde{W}},
\end{equation}

\noindent where $\mathbf{\tilde{W}}$ consists of the first $d$ columns of $\mathbf{W}$ and is usually termed the compression matrix. Consequently, for suitable datasets, most of the information can be retained in a greatly compressed form $\mathbf{\tilde{T}}$. Conversely, the original data matrix $\mathbf{X}$ can be approximately reconstructed by

\begin{equation}
\label{eq:2.4}
\mathbf{\tilde{X}} = \mathbf{\tilde{T}}\mathbf{\tilde{W}}^\intercal,
\end{equation}

\noindent where $\mathbf{\tilde{X}} = [\mathbf{\tilde{x}}_1, \mathbf{\tilde{x}}_2, \dots, \mathbf{\tilde{x}}_n]^\intercal$ is the recovered version of $\mathbf{X}$. It should be pointed out that if the number of features is larger than the sample size of the original dataset, \ie~$n<l$, the transformation matrix, $\mathbf{W}$, for the full transformation (with no loss of information) of $\mathbf{X}$ will possess the dimension of $l$ by $n-1$. Such a feature is very helpful for the representation of hull surfaces, which will be discussed in Section~\ref{sec:hull_form}.

\subsection{Hull-form Compression and Reconstruction with PCA}
\label{sec:hull_form}
Naval architects describe the shape of a hull with a table of $(x, y, z)$ coordinates on its surface, with $x$, $y$, and $z$ measured in the longitudinal, lateral, and vertical directions, respectively. The $x$ and $z$ coordinates in the table are given by the intersections of the transverse sections called stations and waterlines~\cite{watson2002}. Then, the $y$ coordinates on the hull surface, which are called the offset values (measured from the centerplane of the hull), are given at the prescribed sets of $x$ and $z$ positions. Consequently, the tabulation documenting all offset values is called the offset table. In practice, the values of $x$, $y$, and $z$ coordinates are usually normalized by the hull length $L$, half-beam width $B/2$, and draft $T$ of the vessel:

\begin{equation}
\label{eq:2.5}
x^* = \frac{x}{L},\;\; y^* = \frac{y}{B/2},\;\ z^* = \frac{z}{T}.
\end{equation}

Therefore, to represent and compare the hull forms of different vessels, a predetermined grid of stations and waterlines, \ie~the same set of $(x^*, z^*)$ coordinates, can be used with a different set of $y^*$ values for each hull shape. However, a large number of offset values are usually required to fully represent the complex hull geometry, which makes the analysis of the shapes of a group of hulls rather cumbersome. Therefore, if the offset table of each hull is reshaped as a column vector sample $\mathbf{y^*}$, and the offset values are taken as features, PCA can be used to compress the data matrix of the offset tables of a group of hulls, so that a handful of principal scores will suffice to represent the geometry of each hull for further evaluations. To achieve this, the normalized offset table of each hull should be zero centered first by subtracting each offset value with the mean value across samples, \ie~for a group of $n$ hulls each with $l$ offset values, the $j$-th offset value of the $i$-th hull, $y^*_{i,j}$, is shifted as

\begin{equation}
\label{eq:2.6}
\overline{y}_{i,j} = y^*_{i,j} - \mu_j
\end{equation}

\noindent where $\mu_j$ is the mean of the $j$-th offset value for the same $(x^*_j, z^*_j)$ coordinate across all $n$ hulls:

\begin{equation}
\label{eq:2.7}
\mu_j = \frac{1}{n}\sum^n_{i=1} y^*_{i,j}.
\end{equation}

\noindent Consequently, the data matrix consisting of normalized, zero-centered offset tables of $n$ hulls, $\mathbf{\overline{Y}}\in\mathbb{R}^{n,l}$, can be obtained from

\begin{equation}
\label{eq:2.8}
\mathbf{\overline{Y}} = [\mathbf{\overline{y}}_1, \ldots , \mathbf{\overline{y}}_n]^\intercal,
\end{equation}

\noindent where $\mathbf{\overline{y}}_i\in\mathbb{R}^{l}$ is formed by arranging the offset values $\overline{y}_{i,j}$ of the $i$-th hull into a column vector. The transformation matrix $\mathbf{\tilde{W}}\in\mathbb{R}^{l,d}$ can be introduced to map $\mathbf{\overline{Y}}$ into a set of principal scores, $\mathbf{\Lambda}\in\mathbb{R}^{n,d}$ by

\begin{equation}
\label{eq:2.9}
\mathbf{\Lambda} = [\bm{\lambda}_1,\ldots,\bm{\lambda}_n]^\intercal = \mathbf{\overline{Y}}\mathbf{\tilde{W}},
\end{equation}

\noindent where $\bm{\lambda}_i = [\lambda_{i,1},...,\lambda_{i,d}]^\intercal$ is the column vector consisting of $d$ principal scores of the $i$-th hull, and $\lambda_{i,j}$ represents the $j$-th principal score of the $i$-th sample hull. Conversely, the original data matrix of zero-centered offset values can be recovered by 

\begin{equation}
\label{eq:2.10}
\mathbf{\overline{Y}} = \mathbf{\Lambda}\mathbf{\tilde{W}}^\intercal.
\end{equation} 

\noindent Furthermore, the number of hulls in the sample set is likely to be much smaller than that of the offset values, \ie~data matrix $\mathbf{\overline{Y}}$ carries much more features than samples with $l>n$. Therefore, only $d = n-1$ principal scores are needed, \ie~only the first $n-1$ columns of $\mathbf{W}$ are retained in $\mathbf{\tilde{W}}$, to fully represent the surface of each hull in the group with no loss of information. 


It should be mentioned that in traditional ship-design practice, similar low-dimensional characterization of hull forms are usually provided by several geometric coefficients with clear geometrical significance, such as the block coefficient $C_B$ and the prismatic coefficient $C_P$ defined as

\begin{equation}
\label{eq:2.11}
C_B = \frac{\Vol}{LBT},\;\; C_P = \frac{\Vol}{A_M L},
\end{equation} 

\noindent where $\Vol$ is the displaced volume, $L$, $B$, and $T$ are the length, beam, and draft of the vessel, respectively, and $A_M$ is the area of the midship section. Ships with fuller hull forms will generally posses larger $C_B$, while $C_P$ is related to the distributions of displacement along the length of the hull. It is discovered that the values of $C_B$ and $C_P$ exert substantial influences on the performances of a vessel; therefore, much previous effort has been invested in studying the relation between ship performances and geometric coefficients by regression analysis~\cite{holtrop1982}, so that the performances of a hull form can be easily estimated without the knowledge of detailed offset values. This approach allows the general dimensions and shape of a vessel to be determined at the early stage of the designing process. 

However, the geometric coefficients do not account for the detailed geometric features of a vessel, and two distinct vessels can have an identical set of $C_B$ and $C_P$. Consequently, the utilization of geometric coefficients can neither fully capture the complexity of hull geometry nor be reliably used to distinguish different hull forms, and the performance predictions based on geometric coefficient may be less reliable as a result. In comparison, PCA not only provides low-dimensionality representations of hull forms for ease of analysis, but also ensures accurate and unique descriptions of different hull forms using principal scores.

\subsection{Generation of New Hull Forms with Principal Scores}
In addition to compressing the geometric representation of existing vessels, PCA can be further utilized to generate new hull forms by adjusting the principal scores identified from the parent hulls. With the compression matrix $\mathbf{\tilde{W}}$ determined from the offset-value data matrix $\mathbf{\overline{Y}}^P$ of the $n$ parent hulls, any vector of principal scores can be transformed into an offset table by Eq.~\eqref{eq:2.10}, and the generated hull form can be viewed as a linear combination of the parent hulls. However, in order for the generated hull form to be realistic, certain limits to the principal values need to be imposed. In the present study, to randomly generate a vector $\bm{\lambda}$ of principal scores, each element $\lambda_j$ is allowed to vary within the limits prescribed by the parent hulls, \ie

\begin{equation}
\label{eq:2.13}
\min_{1\leq i\leq n}(\lambda_{i,j}^P) \leq \lambda_j \leq \max_{1\leq i\leq n}(\lambda_{i,j}^P),\;\; 1\leq j\leq d
\end{equation}

\noindent where $\lambda_{i,j}^P$ is the $j$-th principal score of the $i$-th parent hull. The column vector $\bm{\lambda}$ is then substituted into Eq.~\eqref{eq:2.10} to obtain the offset table, and together with the predefined stations, waterlines, length, beam, and draft, a new hull geometry is conveniently created without directly manipulating the large number of offset values. Furthermore, the geometric features of the different parent hulls can be properly integrated into the new design. 

\begin{figure}[!t]
\centering
\begin{subfigure}{0.495\textwidth}
	\includegraphics[width=1\columnwidth]{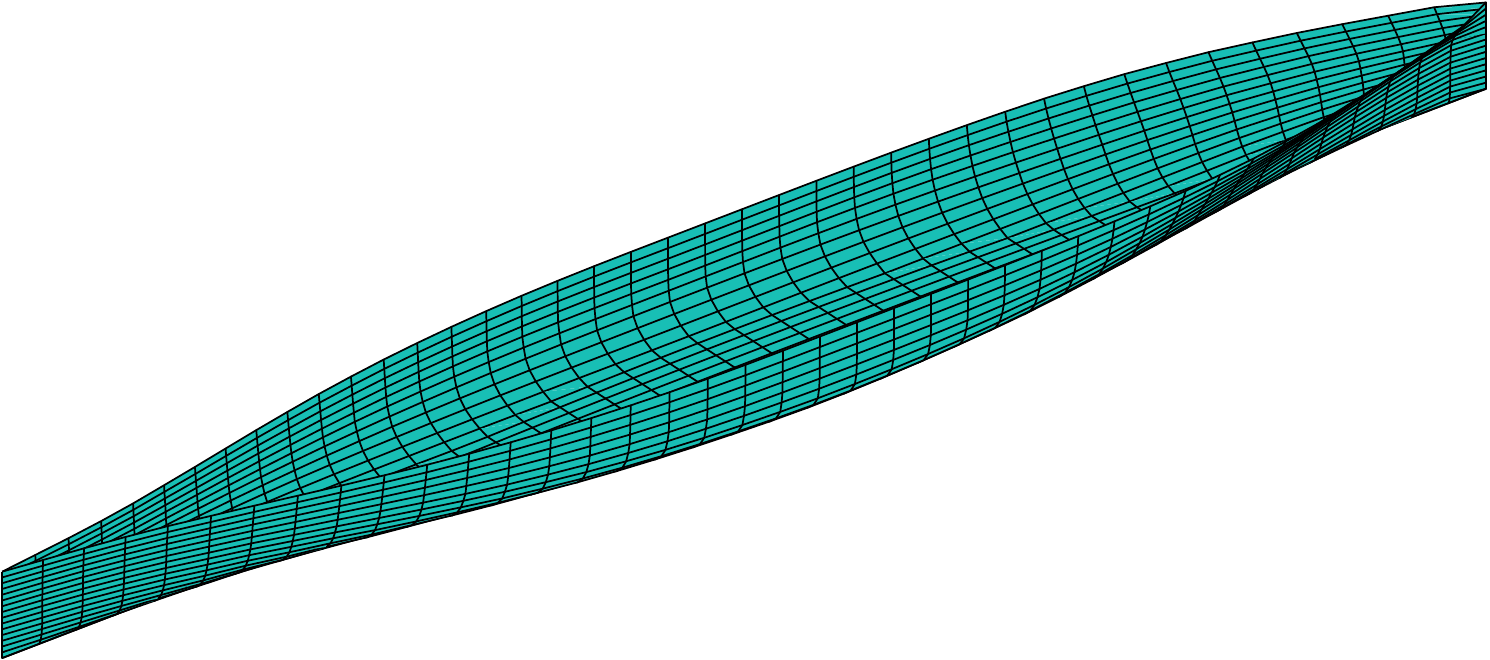}
	\caption{Series 60 $C_B = 0.6$}
	\label{fig:2.1a}
\end{subfigure}
\begin{subfigure}{0.495\textwidth}
	\includegraphics[width=1\columnwidth]{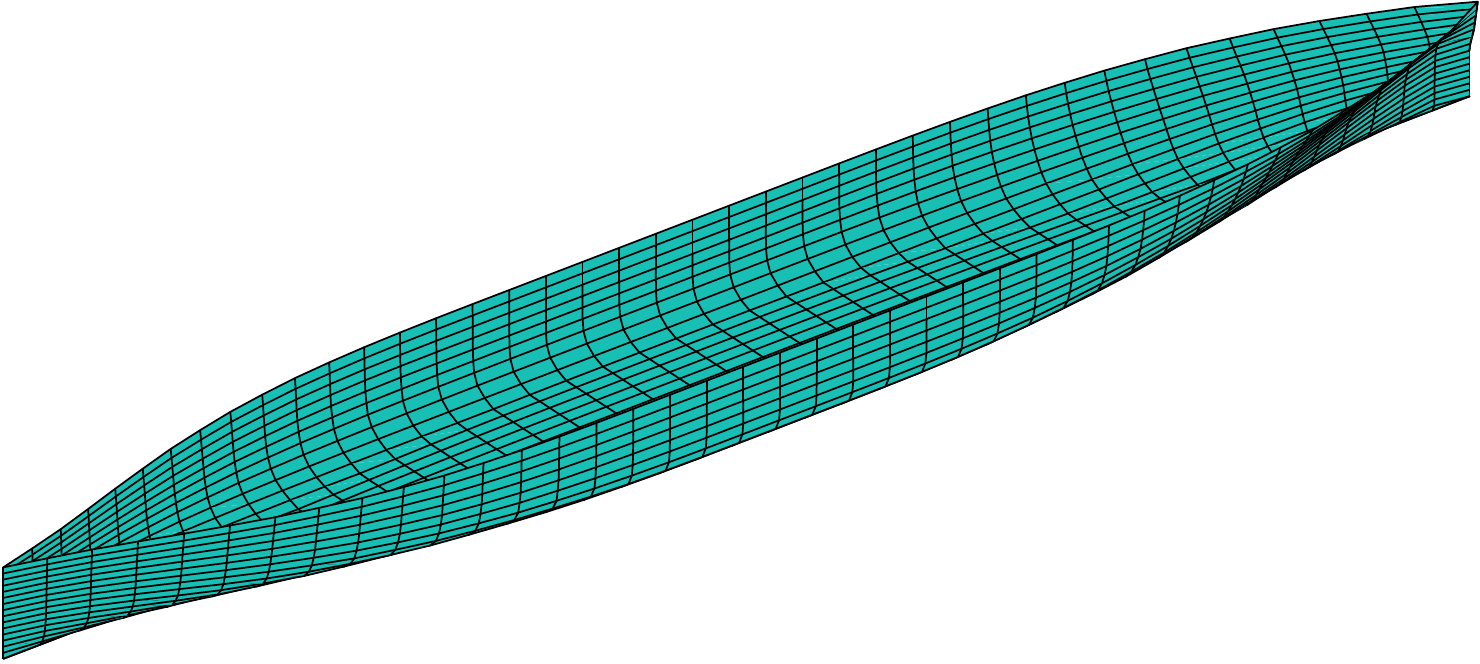}
	\caption{Series 60 $C_B = 0.7$}
	\label{fig:2.1b}
\end{subfigure}
\begin{subfigure}{0.495\textwidth}
	\includegraphics[width=1\columnwidth]{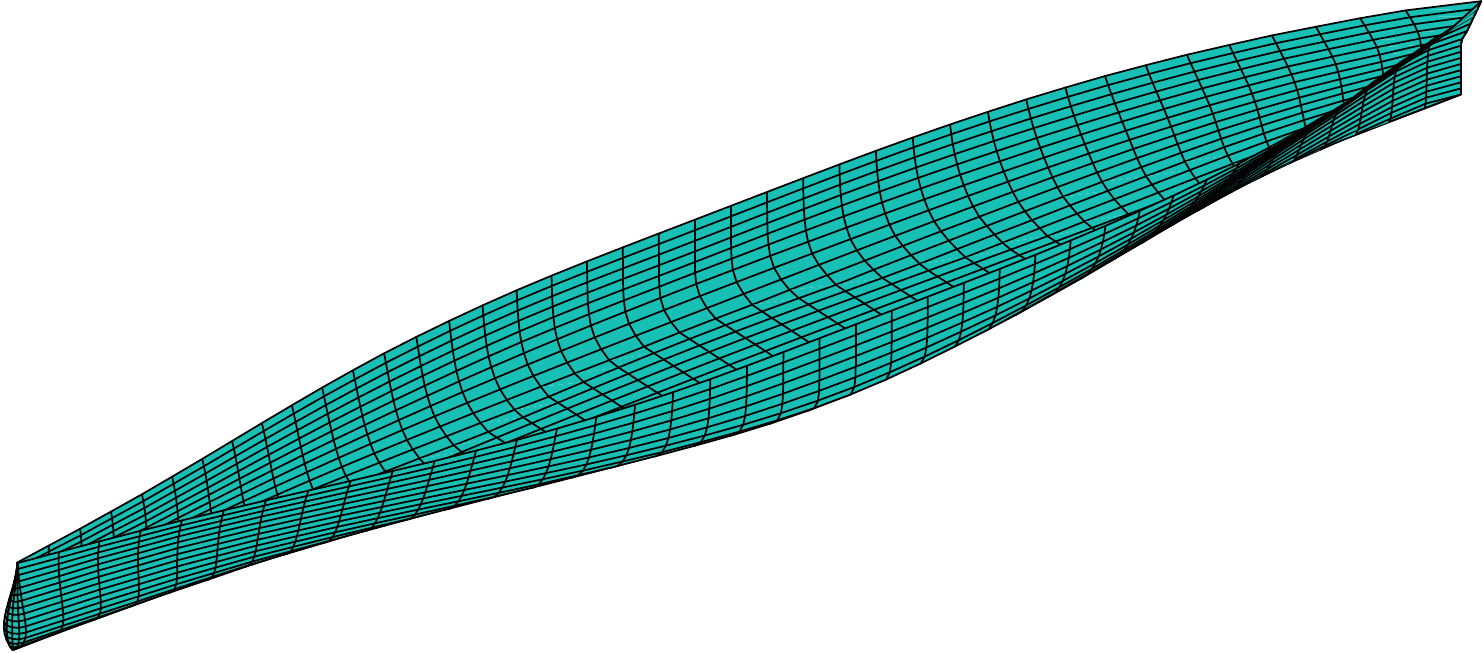}
	\caption{S175 Container ship}
	\label{fig:2.1c}
\end{subfigure}
\begin{subfigure}{0.495\textwidth}
	\includegraphics[width=1\columnwidth]{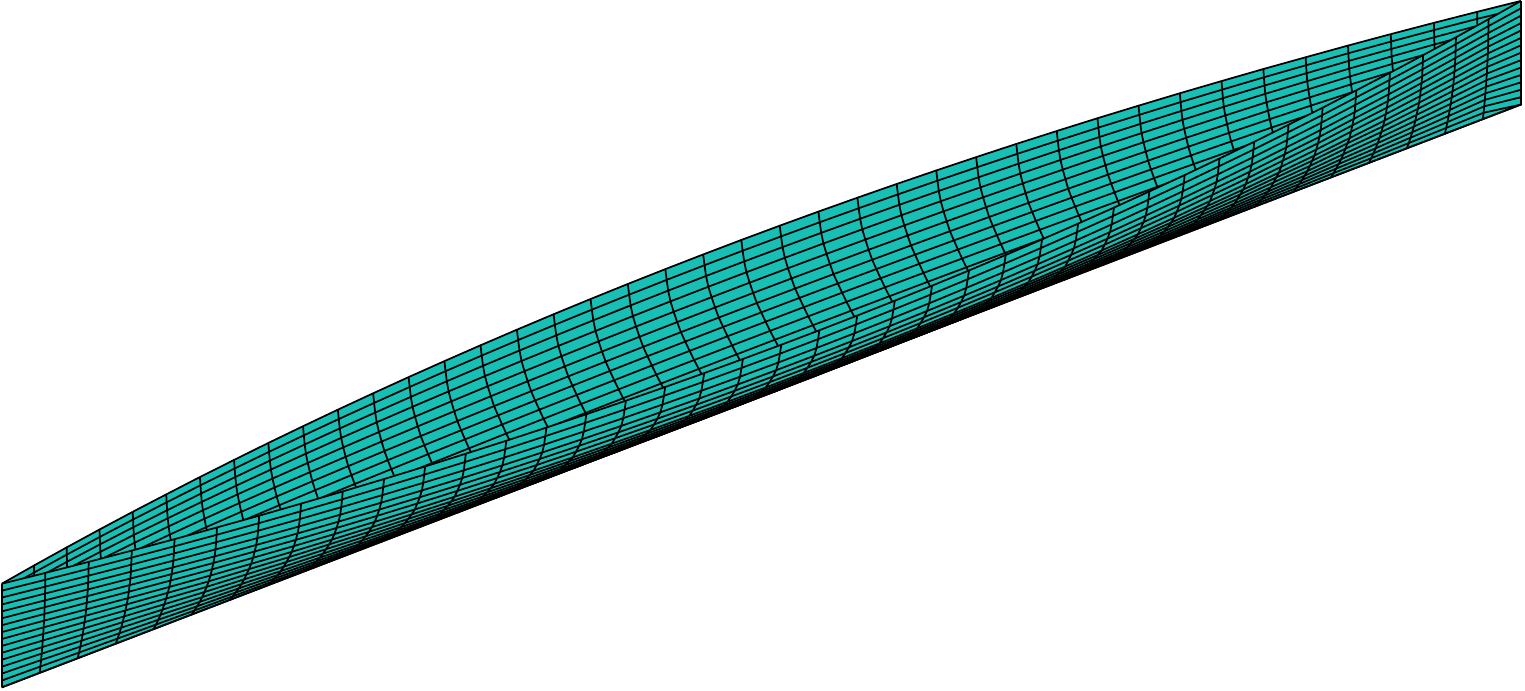}
	\caption{Wigley hull}
	\label{fig:2.1d}
\end{subfigure}
\caption{Three-dimensional views of parent hull forms.}
\label{fig:2.1}
\end{figure}

In the present study, the parent hull forms are provided by four vessels with publicly available offset tables: two Series 60 hulls~\cite{todd1963} with different block efficients ($C_B=0.6$ and 0.7), S175 container ship~\cite{xia1998}, and the Wigley hull. The main dimensions and geometric parameters are listed in Table~\ref{table:2.1}, while the three-dimensional views of the parent hulls are provided in Fig~.\ref{fig:2.1}. The offset table of each of the four parent hull forms consists of 800 offset values, which are taken from the same structured grid made by forty equally-spaced stations and twenty waterlines. Then, a transformation matrix $\mathbf{\tilde{W}}$ is determined by SVD to map the offset values of each parent hull into a vector of three principal scores. In addition, based on the largest and smallest principal scores of the parent hulls, each principal score is scaled to the range of $[0,1]$ by

\begin{equation}
\label{eq:2.14}
\overline{\lambda_{j}} = \frac{\lambda_{j} - \min_{1\leq i\leq 4}(\lambda_{i,j}^P)}{\max_{1\leq i\leq 4}(\lambda_{i,j}^P) - \min_{1\leq i\leq 4}(\lambda_{i,j}^P)},\;\; j=1,\ldots,d
\end{equation}

\noindent and the scaled principal scores of all parent hull forms are also provided in Table~\ref{table:2.1}. It is discovered that the three principal dimensions discovered by PCA account for 85.52$\%$, 13.83$\%$, and 0.65$\%$ of the total variance of the offset values of the parent hulls, respectively. Therefore, the differences among the four parent hulls are mostly accounted for by the first two principal scores.

\begin{table}[!t]
\centering
\caption{Main geometric parameters and principal scores of the parent hulls.}
\begin{tabular}{|c|c|c|c|c|c|c|c|c|}
\hline 
Hull type & $C_B$ & $C_P$ & $L/\Vol^{1/3}$ & $L/B$ & $B/T$ & $\overline{\lambda_{1}}$ & $\overline{\lambda_{2}}$ & $\overline{\lambda_{3}}$ \\ 
\hline 
Series 60 (4210W) & 0.60 & 0.614 & 6.167 & 7.5 & 2.50 & 0.741 & 0.184 & 1.000 \\ 
\hline 
Series 60 (4212W) & 0.70 & 0.710 & 5.593 & 7 & 2.5 & 1.000 & 1.000 & 0.346 \\ 
\hline 
S175 Container Ship & 0.564 & 0.584 & 6.8 & 6.9 & 2.67 & 0.676 & 0 & 0 \\ 
\hline 
Wigley Hull & 0.444 & 0.667 & 7.114 & 10 & 1.6 & 0 & 0.696 & 0.453 \\ 
\hline 
\end{tabular} 
\label{table:2.1}
\end{table}

\begin{figure}[!htbp]
\centering
\begin{subfigure}{0.495\textwidth}
	\includegraphics[width=1\columnwidth]{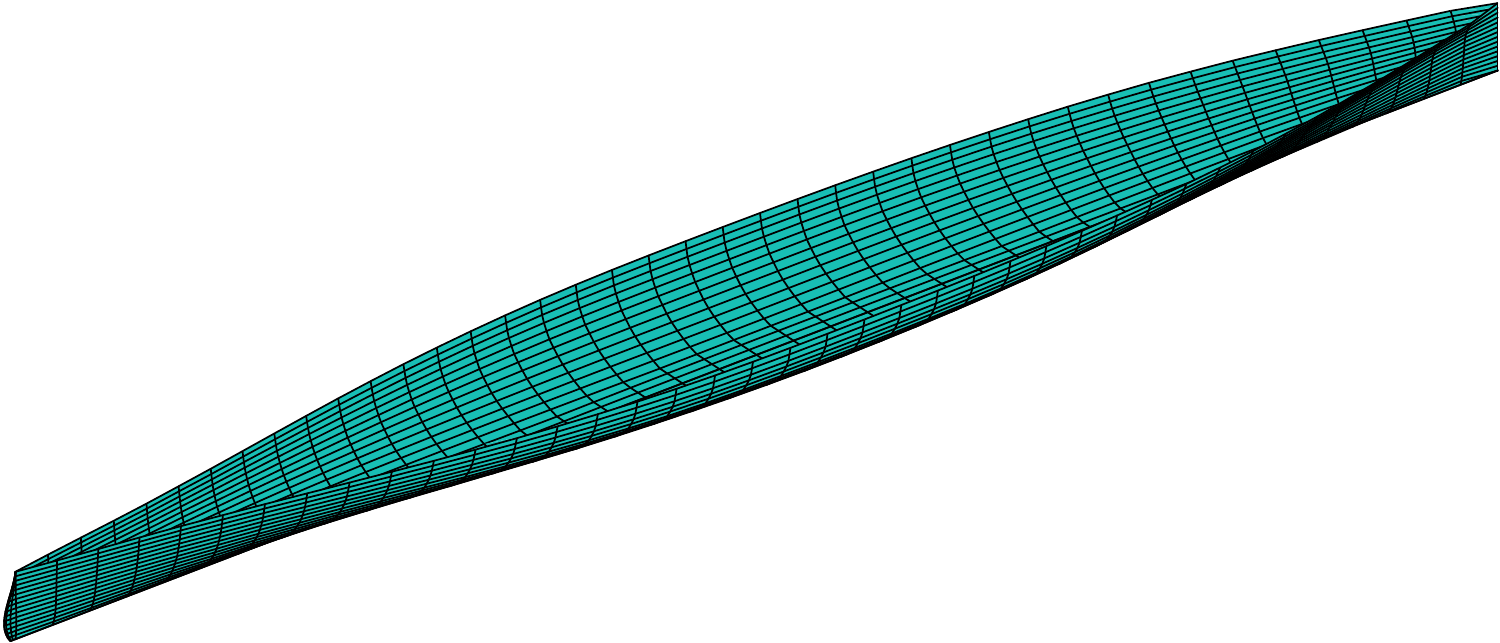}
	\caption{$\bm{\overline{\lambda}} = [0.3,0.2,0.8]$, $L/B=8$, $B/T=3$}
	\label{fig:2.2a}
\end{subfigure}
\begin{subfigure}{0.495\textwidth}
	\includegraphics[width=1\columnwidth]{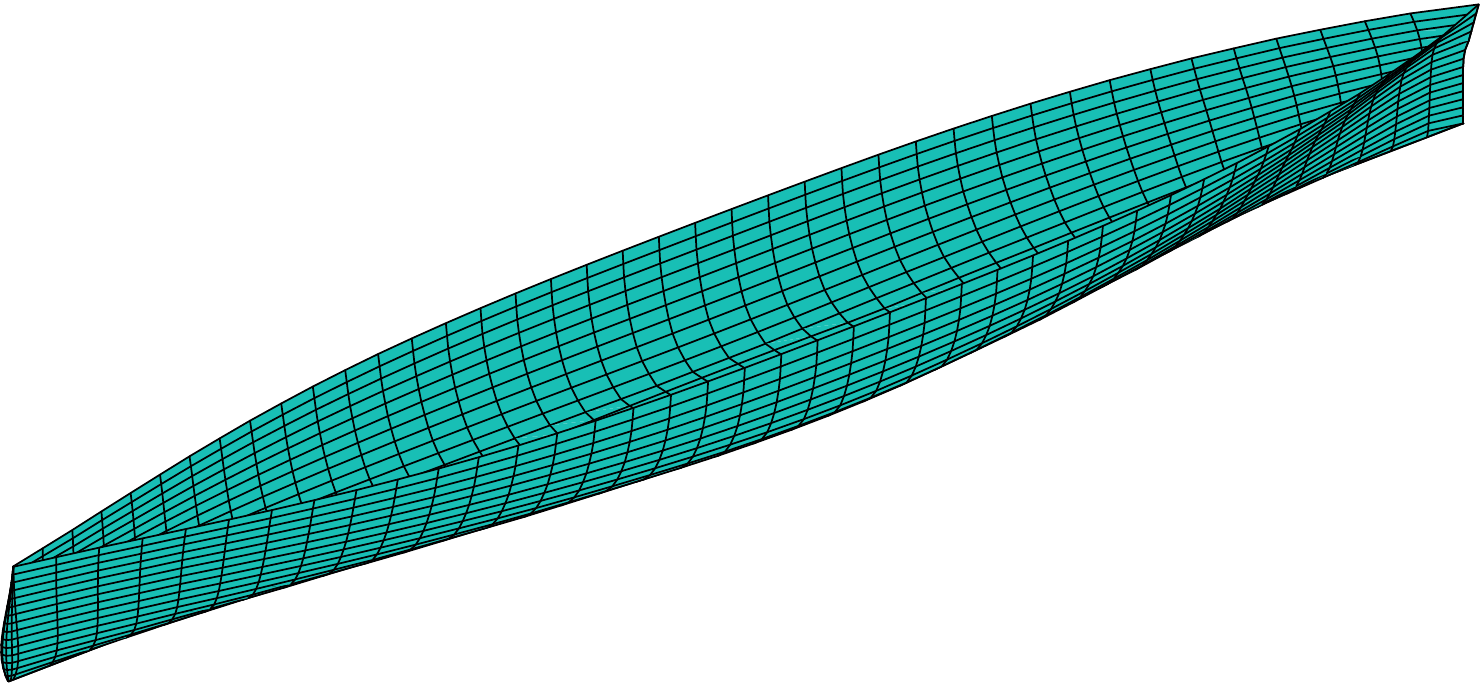}
	\caption{$\bm{\overline{\lambda}} = [0.6,0.5,0]$, $L/B=7$, $B/T=2$}
	\label{fig:2.2b}
\end{subfigure}
\caption{Three-dimensional views of sample generated hull forms.}
\label{fig:2.2}
\end{figure}

Finally, to generate a hull form, one only needs to supply the length of the vessel and the five hull-form parameters consisting of the three principal scores, length-to-beam ratio $L/B$, and beam-to-draft ratio $B/T$. Two of the generated vessels with randomly selected parameters are shown in Fig.~\ref{fig:2.2}. A total of 1131 different hull forms are generated in this manner for evaluation, and the limits for each hull-form parameter are listed in Table.~\ref{table:2.2}, while the length of the vessel is randomly chosen between 150 and 350 meters.

\begin{table}[!htbp]
	\centering
	\caption{Lower and upper limits of each hull-form parameters.}
	\begin{tabular}{|c|c|c|c|c|c|}
		\hline 
		Parameters & $\overline{\lambda}_1$ & $\overline{\lambda}_2$ & $\overline{\lambda}_3$ & $L/B$ & $B/T$  \\ 
		\hline 
		Minimum & 0 & 0 & 0 & 6.9 & 2.0  \\ 
		\hline 
		Maximum & 1 & 1 & 1 & 9 & 3.5 \\ 
		\hline 
	\end{tabular}  	
	\label{table:2.2}
\end{table}

\begin{figure}[!htbp]
\centerline{
\begin{subfigure}{0.5\textwidth}
	\includegraphics[width=1\columnwidth]{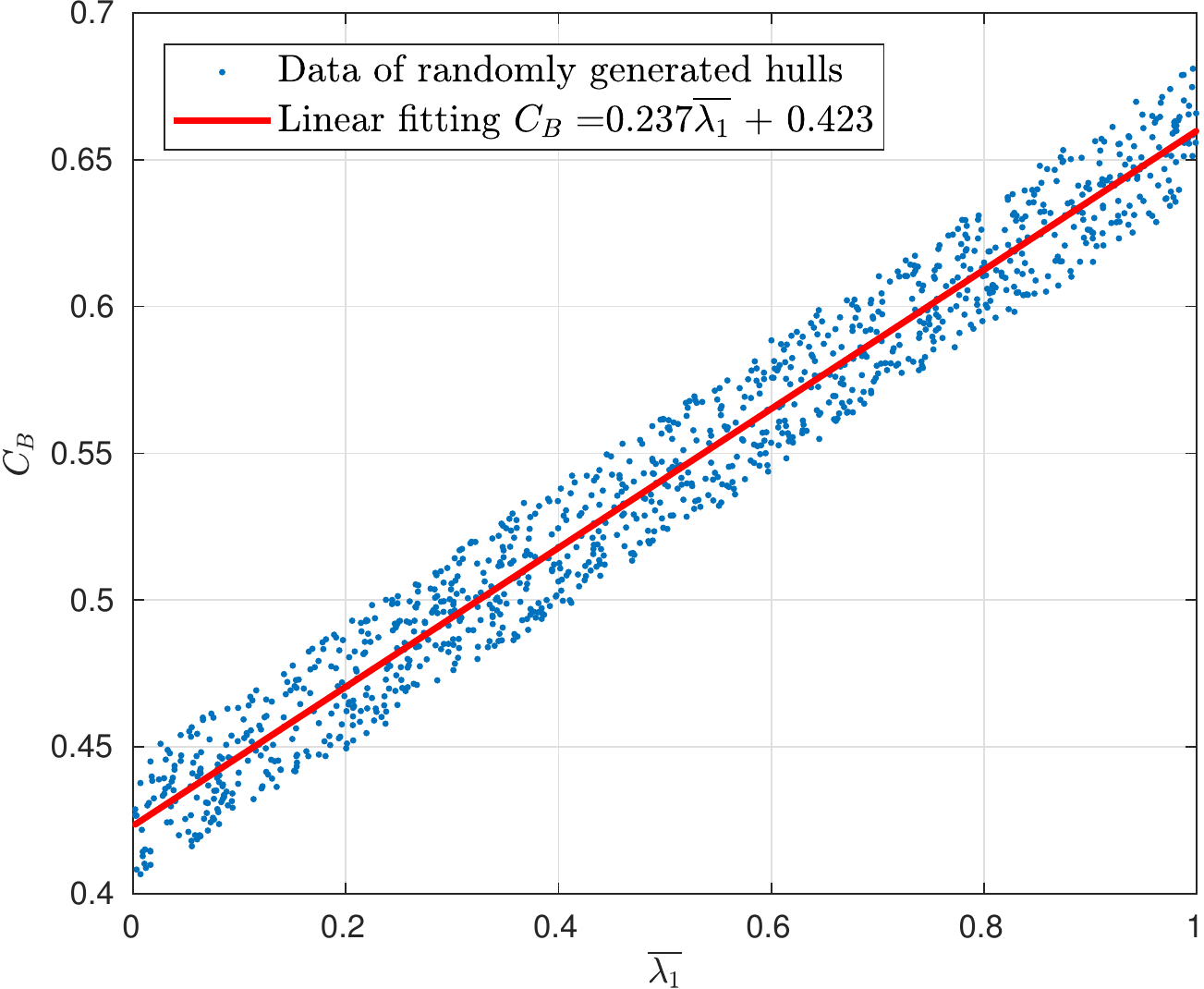}
	\caption{$C_B$ plotted against $\overline{\lambda_1}$}
	\label{fig:2.3a}
\end{subfigure}
\begin{subfigure}{0.5\textwidth}
	\includegraphics[width=1\columnwidth]{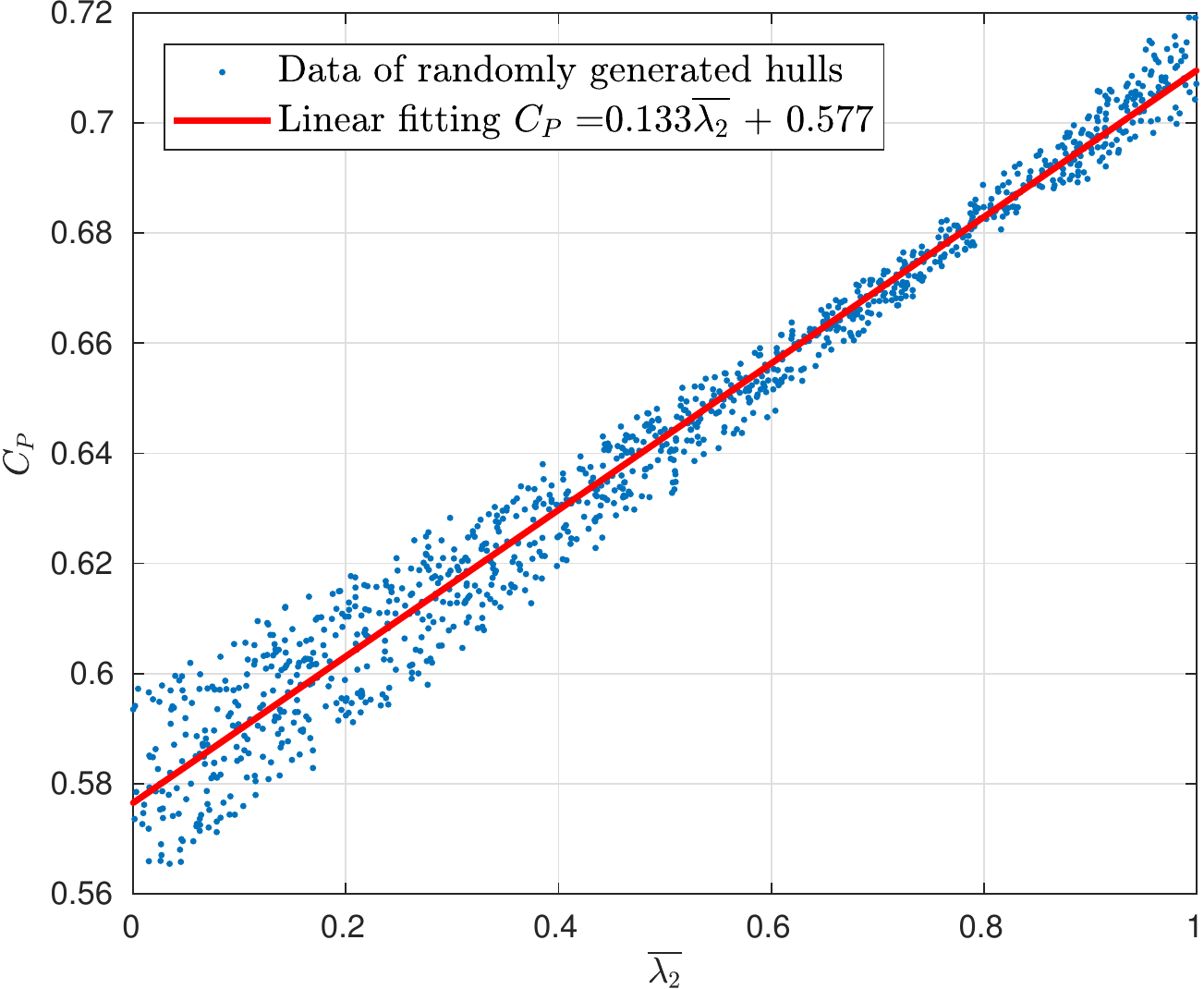}
	\caption{$C_P$ plotted against $\overline{\lambda_2}$}
	\label{fig:2.3b}
\end{subfigure}}
\caption{Selected principal scores plotted against geometric coefficients.}
\label{fig:2.3}
\end{figure}

Although generally it can be difficult to interpret the physical significance of principal scores, some interesting comparisons can be drawn with geometric coefficients. To this end, the values of $\overline{\lambda_1}$ and $\overline{\lambda_2}$ are plotted against those of $C_B$ and $C_P$ of the randomly generated hulls in Fig.~\ref{fig:2.3}. It is discovered that the values of $\overline{\lambda_1}$ and $\overline{\lambda_2}$ linearly correlate with the values of $C_B$ and $C_P$. Therefore, geometric coefficients indeed carry key information about the geometries of hulls because they are highly correlated with the leading principal scores. However, the non-negligible scattering of the data points around the linearly fitted lines also suggests that geometric coefficients fail to capture some variance in the geometric information, and the mapping between offset tables of vessels and their corresponding geometric coefficients is not unique. In essence, both PCA and coefficient-based hull-form representation project the offset values of a hull form into spaces of much lower dimensionality. However, PCA provides a more systematic approach to generate parameters that are orthogonal and in the directions of maximum possible variance with the given selection of hull geometries. If a larger data base of parent hull forms is provided, more PCA parameters can be introduced to cover minor design features of the hull, such as bulbs bows, that are not properly represented by conventional geometric coefficients.

\section{Computational Evaluation of Vessel Calm-Water Performance}
Resistance experienced by a vessel in calm water is directly related to its efficiency and power requirement; therefore, the minimization of calm-water resistance is frequently one of the goals of hull-form optimization. In the present study, to provide the training and testing samples needed for building a DNN to carry out hull-form optimization, the calm-water resistances of the randomly generated hull forms introduced in the previous sections are computationally evaluated. 

\subsection{Decomposition of Calm-Water Resistance}
The calm-water resistance of a vessel comes from various sources. In accordance with the guidelines provided by the International Towing Tank Conference (ITTC~\cite{ITTC2011}), the total resistance of a vessel in calm water is decomposed into frictional drag $R_F$, residual resistance $R_R$, and minor contributions such as air resistance $R_{air}$ and the drag induced by appendages $R_{app}$~\cite{edward1989, zhang1997}. Frictional drag $R_F$, which is related to the viscosity of water, is computed by

\begin{equation}
\label{eq:3.1}
R_F = \frac{1}{2}\rho U^2 SC_F,
\end{equation}

\noindent where $\rho$ is the density of water, $S$ is the wetted surface area of the ship, and $C_F$ is the coefficient of frictional drag given by the ITTC correlation-line formula~\cite{ITTC1957} as

\begin{equation}
\label{eq:3.2}
C_F = \frac{0.075}{(\log_{10}Re - 2)^2},
\end{equation}

\noindent in which $Re = UL/\nu$ is the Reynolds number and $\nu$ is the kinematic viscosity of water taken as $1.016\times 10^{-6} \text{m}^2/s$ in the present study. 

Meanwhile, residual resistance $R_R$ mainly consists of the wave resistance $R_W$, which accounts for the loss of energy through the generation of ship waves, and a small correction factor is applied to the value of $R_W$ to account for the form drag resulting from flow separation due to the curvature of the hull surface~\cite{min2010}, \ie

\begin{equation}
\label{eq:3.3}
R_R = (1+k)R_W,
\end{equation}

\noindent where $k$ is the form factor. However, hull forms investigated in the present study are generally slender; therefore, the contribution from form drag is negligible with $k\approx 0$. Similarly, the contributions of appendages and air resistance are also relatively small and neglected in the present study. Consequently, the total resistance $R_T$ is approximated as

\begin{equation}
\label{eq:3.4}
R_T = R_F + (1+k)R_W + R_{air} + R_{app} \approx R_F + R_W.
\end{equation}

In addition, the forward speed of a vessel, which strongly influences the calm-water resistance, is represented in dimensionless form by the Froude number:

\begin{equation}
\label{eq:3.7}
F_n = \frac{U}{\sqrt{gL}}.
\end{equation}

\noindent Previous studies and engineering practices show that frictional drag is more significant for vessels traveling with lower Froude numbers~\cite{faltinsen2005}, while wave resistance is dominant at higher forward speeds. Furthermore, to represent the calm-water efficiency of a vessel, the total resistance $R_T$ is normalized by the vessel displacement:

\begin{equation}
	\label{eq:3.5}
	\overline{C}_T = \frac{R_T}{\rho g \Vol},
\end{equation} 

\noindent where $\overline{C}_T$ is the merit coefficient, which is in fact the cost-to-payload ratio of the vessel. It is not hard to deduce that for a given forward speed, vessels with smaller $\overline{C}_T$ achieves better fuel efficiency by requiring less power to deliver the same amount of payload. However, in practice, ships do not travel at a constant speed, and a more realistic representation of vessel efficiency should consider its entire operational profile~\cite{baldi2015} of all possible operating speeds and their associated distributions. Consequently, a more practical criterion which represents the cumulative cost-to-payload ratio over a range of speed, termed as the operational merit coefficient $\beta$ is defined as
\begin{equation}
\label{eq:3.6}
\beta = \int_{F_{n_l}}^{F_{n_u}} w(F_n)\overline{C}_T \, \mathrm{d}F_n,
\end{equation}

\noindent where $F_{n_l}$ and $F_{n_u}$ are the lower and upper bounds of the operating Froude number, and $w$ is the weight function specifying the distributions of $F_n$ in the operating range. Therefore, from the perspective of transporting efficiency, the optimal hull forms pursued in the present study are the ones with lowest values of $\beta$ for given speed ranges.

\subsection{Computation of Wave Resistance by GPU-accelerated Potential-flow Panel Method}
While the frictional drag of a vessel can be easily and rather accurately acquired from Eq.~\eqref{eq:3.1} and~\eqref{eq:3.2}, the inviscid wave-resistance component  $R_W$ is acquired by solving the potential-flow boundary-value problem of a vessel moving steadily in calm-water~\cite{yeung1982}. 

\begin{figure}[!htbp]
\centering
\includegraphics[width=0.5\columnwidth]{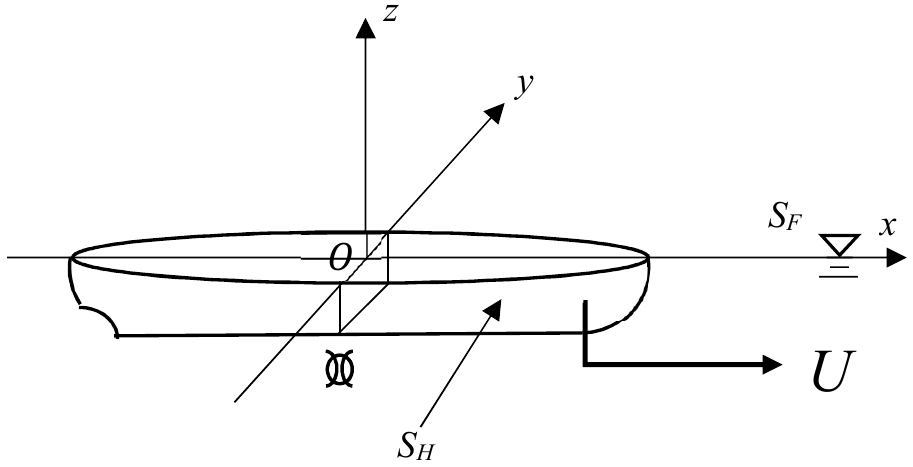}\\
\caption{Problem schematic for computing wave resistance.}
\label{fig:3.1}
\end{figure}

A moving frame of reference $Oxyz$ as depicted in Fig.~\ref{fig:3.1} is fixed on the ship traveling towards the positive $x$-direction at constant speed $U$ in a laterally unbounded fluid domain of infinite depth. The total velocity field of the domain can be represented by the superposition of vessel disturbance and a uniform background flow of speed $U$ in the negative-$x$ direction~\cite{raven1996}. The hull surface and the free surface are denoted as $S_H$ and $S_F$, respectively. The fluid is assumed inviscid, incompressible, and the flow irrotational. The disturbance velocity generated by the presence of the vessel, $\mathbf{u}$, is given by the gradient of the velocity potential $\phi$,

\begin{equation}
\label{eq:3.8}
\nabla\phi = \mathbf{u},
\end{equation}

\noindent Then, the boundary value problem for $\phi$ in the fluid domain is given by

\begin{subequations}
\label{eq:3.9}
\begin{align}
& \nabla^2\phi = 0,\label{eq:3.9a}\\
& \phi_n  = Un_x \quad\text{on $S_H$},\label{eq:3.9b}\\
& \phi_{z} = 0  \quad \text{as}\quad z\rightarrow -\infty,\label{eq:3.9c}\\
& \phi_{xx}=\phi_x=0 \quad \text{as}\quad x\rightarrow +\infty,\label{eq:3.9d}\\
& \phi_x\zeta_x + \phi_y\zeta_y - \phi_z - U\zeta_x = 0\quad\text{on $S_F$},\label{eq:3.9e}\\
& \frac{1}{2}|\nabla\phi|^2 -U\phi_x +g\zeta = 0 \quad\text{on $S_F$}, \label{eq:3.9f}
\end{align}
\end{subequations}

\noindent where $n = (n_{x}, n_{y}, n_{z})$ is the unit interior normal to the hull surfaces $S_{H}$, $\zeta(x,y,t)$ is the free-surface elevation. Eq.~\eqref{eq:3.9a} is the Laplace's equation governing the velocity potential $\phi$, Eq.~\eqref{eq:3.9b} is the non-penetration boundary condition on the hull surface, Eqs.~\eqref{eq:3.9c} and ~\eqref{eq:3.9d} specify the boundary conditions at infinite water depth and far upstream, while ~\eqref{eq:3.9e} and~\eqref{eq:3.9f} are the kinematic and dynamic free-surface boundary conditions observed in the moving frame of reference~\cite{wehausen1973}, respectively. In the present study, with the assumption that the vessel exerts small disturbances, the nonlinear free-surface boundary conditions are expanded around the calm-water surface ($z = 0$), and the terms up to second order in wave slope are kept~\cite{tarafder2007}.

The numerical solution of the above boundary value problem is given by a potential-flow panel method called the Simple-Source Panel Method (SPM)~\cite{yu2017iwwwfb}. In the SPM computation the surfaces enclosing the fluid domain is discretized into flat quadrilateral panels with mixed source and dipole distributions of constant strength~\cite{hess1964}. A linear system of equations is constructed to solve for the disturbance velocity potential $\phi$ on the hull surface and the free surface. Then, the unsteady Euler Integral~\cite{wehausen1960} is invoked to compute the hydrodynamic pressure, which is later integrated over the hull surface to acquire the wave resistance of the vessel. Details regarding the potential-flow panel method is extensively documented in the existing literature (see \eg~\cite{wehausen1960, yeung1982, dawson1977, raven1996, tarafder2007}) and will not be repeated here. 

\begin{figure}[!htbp]
	\centerline{\begin{subfigure}{.53\textwidth}
			\includegraphics[width=1\columnwidth]{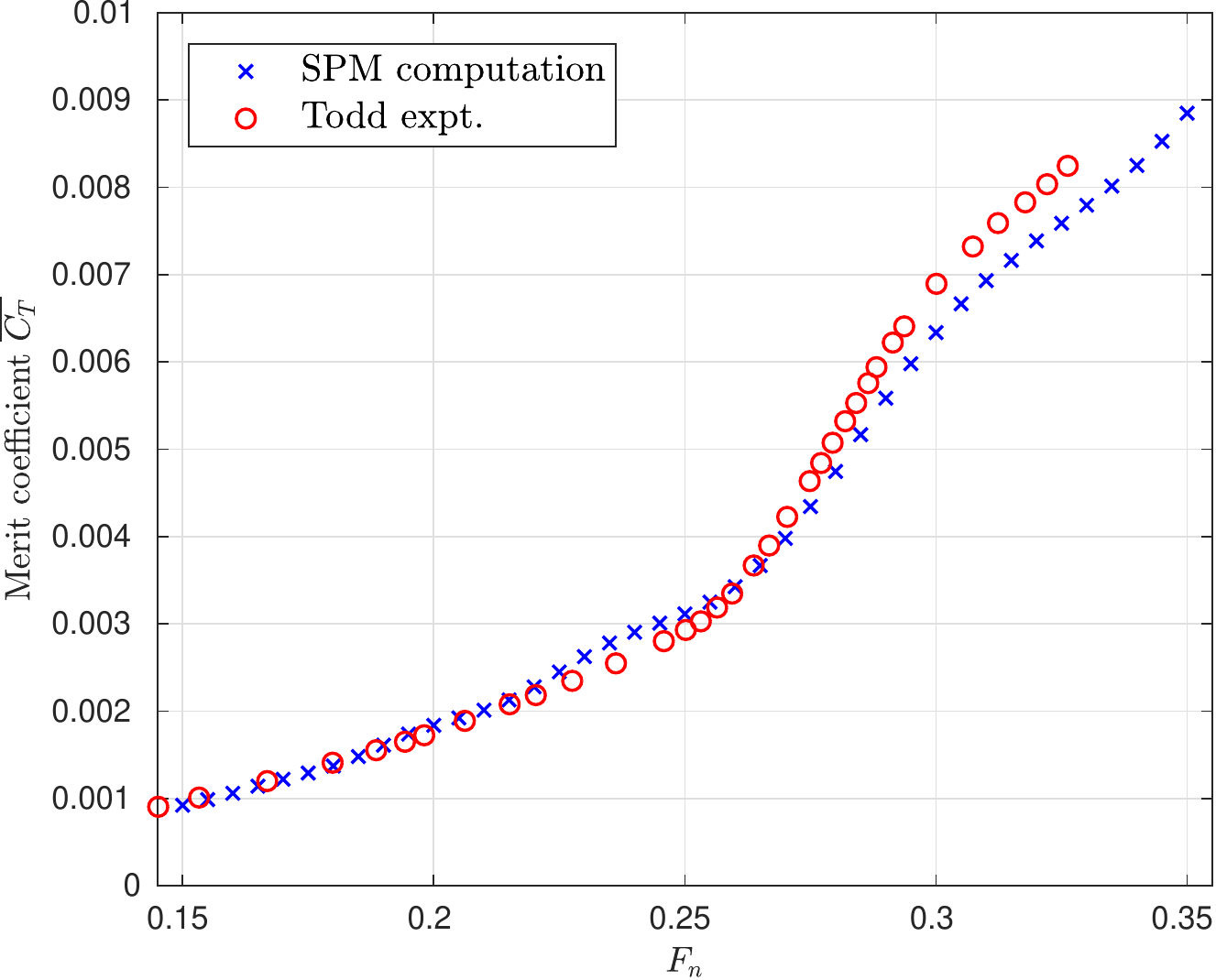}
			\caption{Results for Series 60, $C_B=0.6$.}
			\label{fig:3.2a}
		\end{subfigure}%
		\begin{subfigure}{0.53\textwidth}
			\includegraphics[width=1\columnwidth]{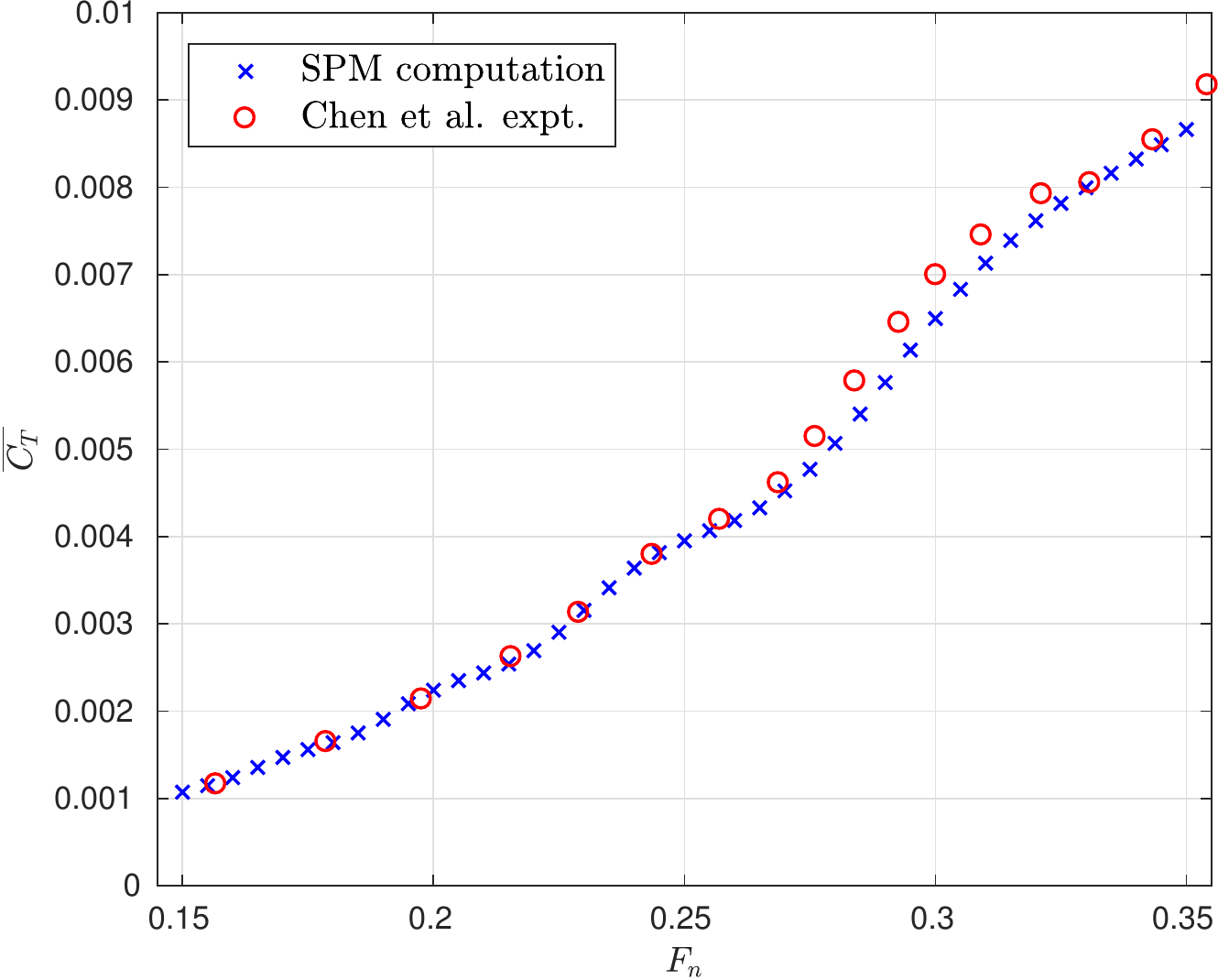}
			\caption{Results for the Wigley hull.}
			\label{fig:3.2b}
		\end{subfigure}}
		{\caption{SPM computational results with comparison to published experimental measurements~\cite{todd1963, chen2016}.}\label{fig:3.2}}
	\end{figure}

Compared to viscous-fluid CFD simulations and towing-tank experiments, potential-flow computation usually provides sufficiently accurate results at much smaller computing costs. Such a feature is desirable for the training of a DNN for hull-form optimization, since the training and testing samples must come with sufficient quantity and satisfactory accuracy.  In addition, the SPM computation is accelerated by a Graphic Processing Unit (GPU) which has exceptional parallel computing capabilities. 

The wave resistances at twenty-one different Froude numbers ranging from 0.15 to 0.35 are computed for all 1131 randomly generated hull forms. The computation was performed on a desktop PC with Advanced Micro Devices 16-core Threadripper CPU running at 3.4Ghz and an Nvidia GTX Titan Black GPU, and the entire computation took less than 48 hours. The viscous skin-friction resistances provided by Eq.~\eqref{eq:3.2} are then added to acquire total resistances, which are further normalized into merit coefficient $\overline{C}_T$. To validate the SPM computation, the computational results of $\overline{C}_T$ for Series 60 hull with $C_B=0.6$ and Wigley hull are compared with published experimental measurements in Fig.~\ref{fig:3.2}, and satisfactory agreement is observed.  

\section{Deep Neural Network (DNN) for Hull-form Optimization}
\subsection{Brief Introduction of DNN with Application to Hull-Form Optimization}
For some function $\mathbf{s} = g(\mathbf{z})$ which maps input $\mathbf{z}$ into output $\mathbf{s}$ by a mapping function $g$, a classic MultiLayer Perceptron (MLP) neural network~\cite{looney1997} can learn to establish the best approximation $\mathbf{s}' = g'(\mathbf{z};\mathbf{\theta})$, by adjusting the parameters $\mathbf{\theta}$ based on pre-labeled examples~\cite{goodfellow2016}. A typical MLP neural network consists of an input layer, hidden layers, and an output layer, and a Deep Neural Network (DNN) refers to MLP neural networks with a very deep structure, \ie~many hidden layers populated with artificial neurons, so that highly complicated relations between inputs and outputs can be better comprehended and approximated.

The training of a DNN starts with submitting inputs $\mathbf{z}$ at the input layer. The inputs then go through linear transformations and processed by the nonlinear activation functions by the neurons in the hidden layers, and the results are passed on to the next hidden layer until the output layer is reached. This process is called the forward propagation. The output of the network, $\mathbf{s}'$, is then compared with the expected outcome $\mathbf{s}$ of the mapping $\mathbf{g}$. Then, through a process referred to as the backward propagation, the weight and bias of each neuron on the hidden layers are adjusted layer by layer backward in the direction of minimizing the differences between $\mathbf{s}$ and $\mathbf{s}'$. The above process is repeated multiple times (``epochs") until satisfactory agreement between $\mathbf{s}$ and $\mathbf{s}'$ is achieved. All pre-labeled examples are typically divided into two sets: a training set for training the DNN and a test set for checking the accuracy of the predictions. We have omitted the use of a separate validation set for tuning hyperparameters in the present work. Details regarding the training, testing, and validation of DNNs can be found in the literature~\cite{lippmann1987, lecun2015, geron2017}. In addition, with the help of GPU and modern machine-learning framework such as TensorFlow~\cite{tensorflow2015-whitepaper} and Keras~\cite{chollet2017}, forward and backward propagations can be executed in a massively parallel fashion on GPUs~\cite{abadi2016}, which makes the training, testing, and application of DNN on large datasets very fast and convenient. 

In the case of hull-form evaluations, if the input $\mathbf{z}$ is taken as the geometric features and forward speed of vessels, and the output $\mathbf{s}$ is taken as the merit coefficients, a DNN may be trained on an adequate amount of training samples provided either through numerical computation or experimental measurements to accurately approximate the mapping from the hull speed and geometry to the associated calm-water performance. Furthermore, since the geometric representation of a hull form is substantially compressed by PCA in advance, the complexity of the network can be greatly reduced as a result of much fewer inputs. Eventually, with the fast, parallel forward propagation on the DNN, large amount of hull forms can be quickly evaluated without invoking the much more expensive direct computational or experimental hydrodynamic analyses, and the optimal hull form, which features a suitable combination of the characteristics of the parent hulls for a given optimizing target can be rapidly identified.
 
\subsection{Training and Testing of the DNN}
As introduced in Section~\ref{sec:2}, the geometries of the hull forms are specified by the five hull-form parameters listed in Table.~\ref{table:2.2}. These parameters, along with the Froude number $F_n$ and Reynolds number $R_e$, which together specify the ship length and speed, are taken as the inputs to the DNN, while the outputs are the associated merit coefficient $\overline{C}_T$ . Out of the 1131 computationally evaluated hull forms, 1006 of them are randomly selected to form the training set, while the rest are assigned to the test set. In addition, since each generated hull form is evaluated at 21 different forward speeds, the training set and test set consist of 21126 and 2625 samples, respectively.
 
\begin{figure}[!t]
\centerline{  
	\includegraphics[width=1\columnwidth]{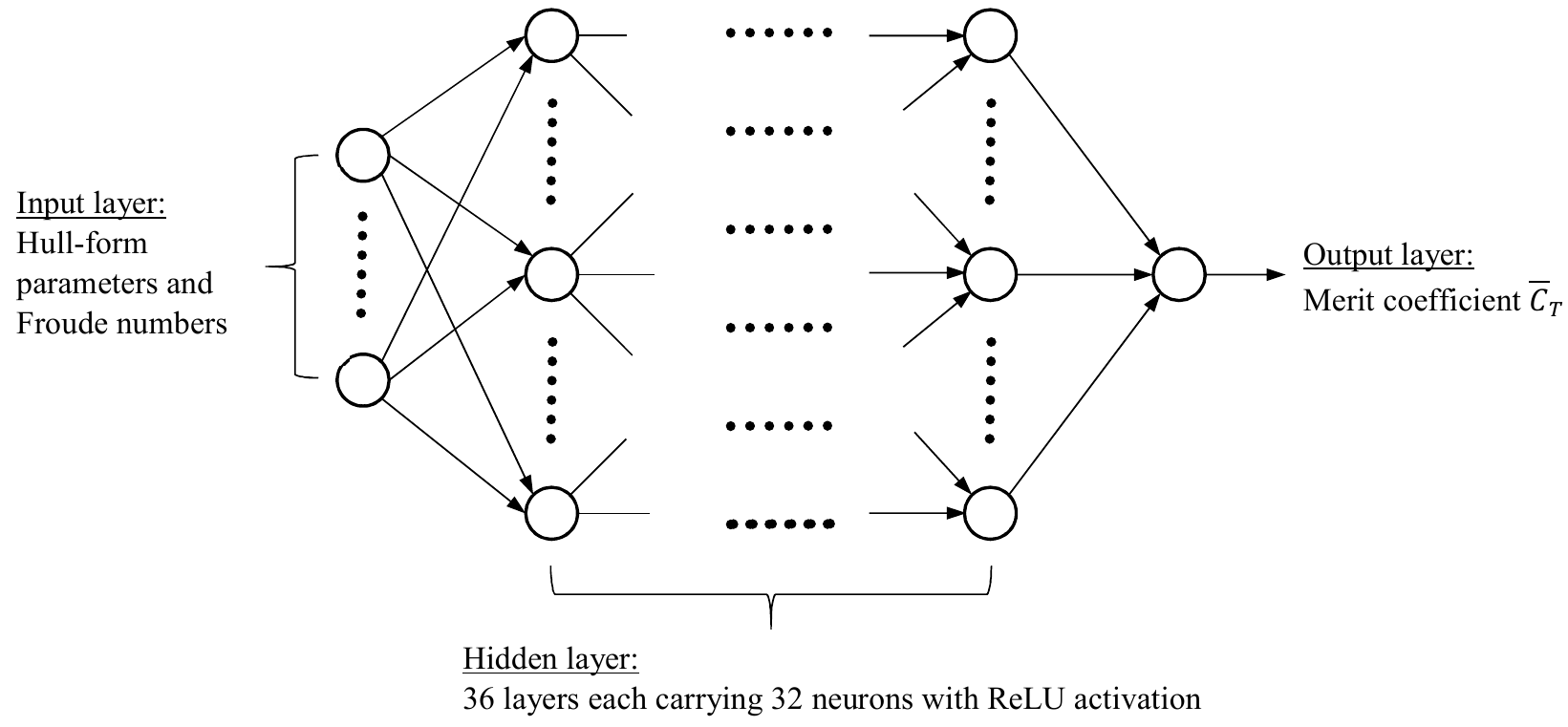} 
       }
    {\caption{Structure of the DNN for hull form evaluations.}
    \label{fig:4.1}}
\end{figure}

The DNN used in the present study consists of 36 hidden layers with each layer populated by 32 neurons, as shown in Fig.~\ref{fig:4.1}. The Rectified Linear Unit (ReLU)~\cite{nair2010} is selected as the activation function for all neurons. At the training stage, the network is initialized with the method introduced in He \etal~\cite{he2015} and optimized with Adaptive Moment Estimation (Adam)~\cite{kingma2014} to minimize the mean absolute percentage error (MAPE) $\delta$~\cite{tofallis2015}:

\begin{equation}
\label{eq:4.1}
\delta = \frac{1}{m}\sum^m_{i=1}\lvert\frac{\overline{C}_{T_i} - \hat{\overline{C}}_{T_i}}{\overline{C}_{T_i} }\rvert \times 100 \%,
\end{equation}
where $\hat{\overline{C}}_T$ is the DNN prediction of merit coefficient, and $m$ is the number of training samples. The hyper-parameters for the Adam optimizer are kept the same as introduced in~\cite{kingma2014}. The Keras machine-learning framework with TensorFlow backend is utilized for the training and testing of the DNN and the ensuing hull-form optimization. The network is trained with 8000 epochs with a batch size of 2000 on an Nvidia GTX 1080 Ti GPU. At the training stage of the DNN, the test error at the end of each epoch is monitored, and the network with the lowest test error, which is $0.942\%$, is saved and utilized for the ensuing hull-form evaluation and optimization.

\begin{figure}[!htbp]
\centerline{\begin{subfigure}{.53\textwidth}
        \includegraphics[width=1\columnwidth]{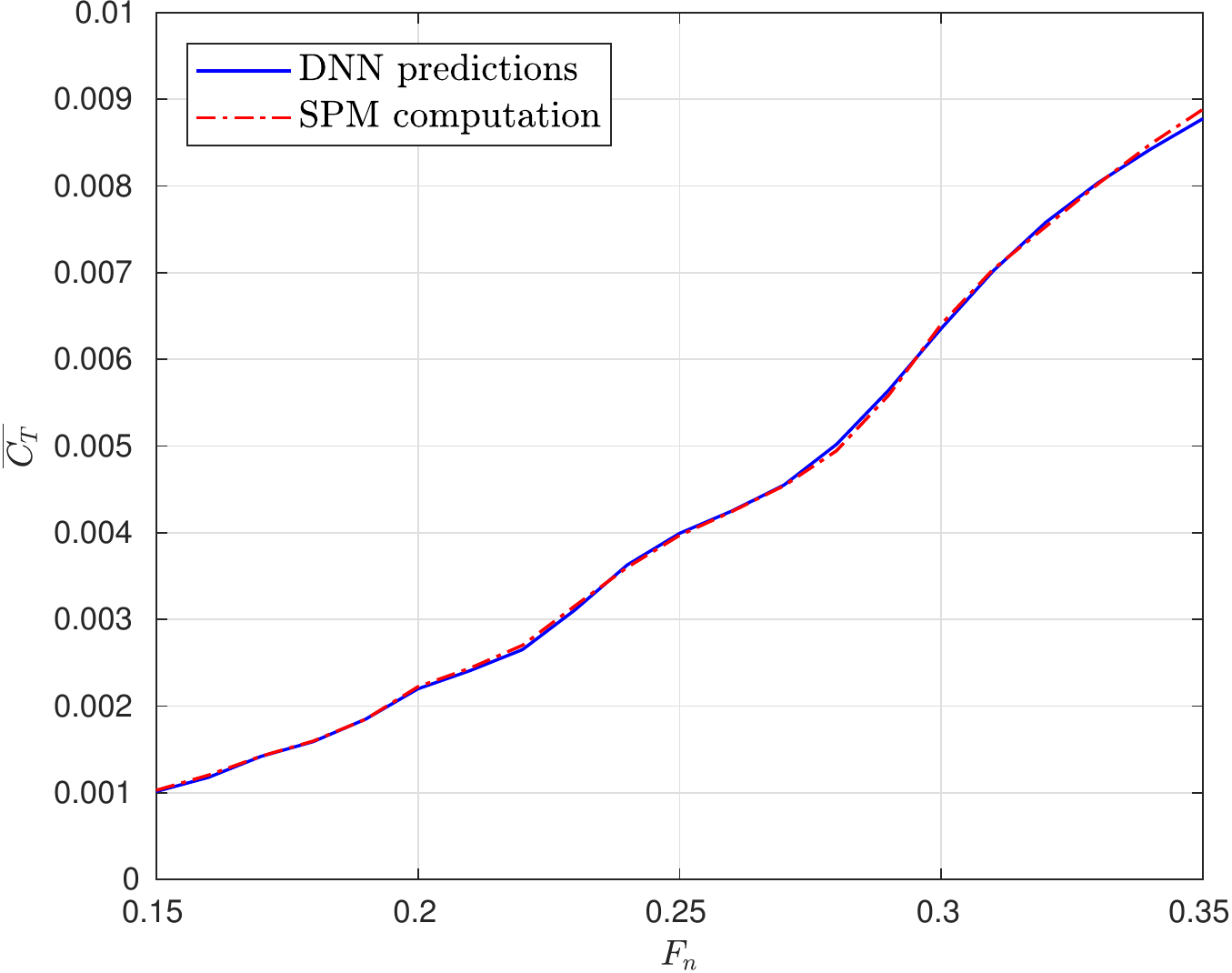}
        \caption{}
        \label{fig:4.2a}
    \end{subfigure}%
    \begin{subfigure}{0.53\textwidth}
        \includegraphics[width=1\columnwidth]{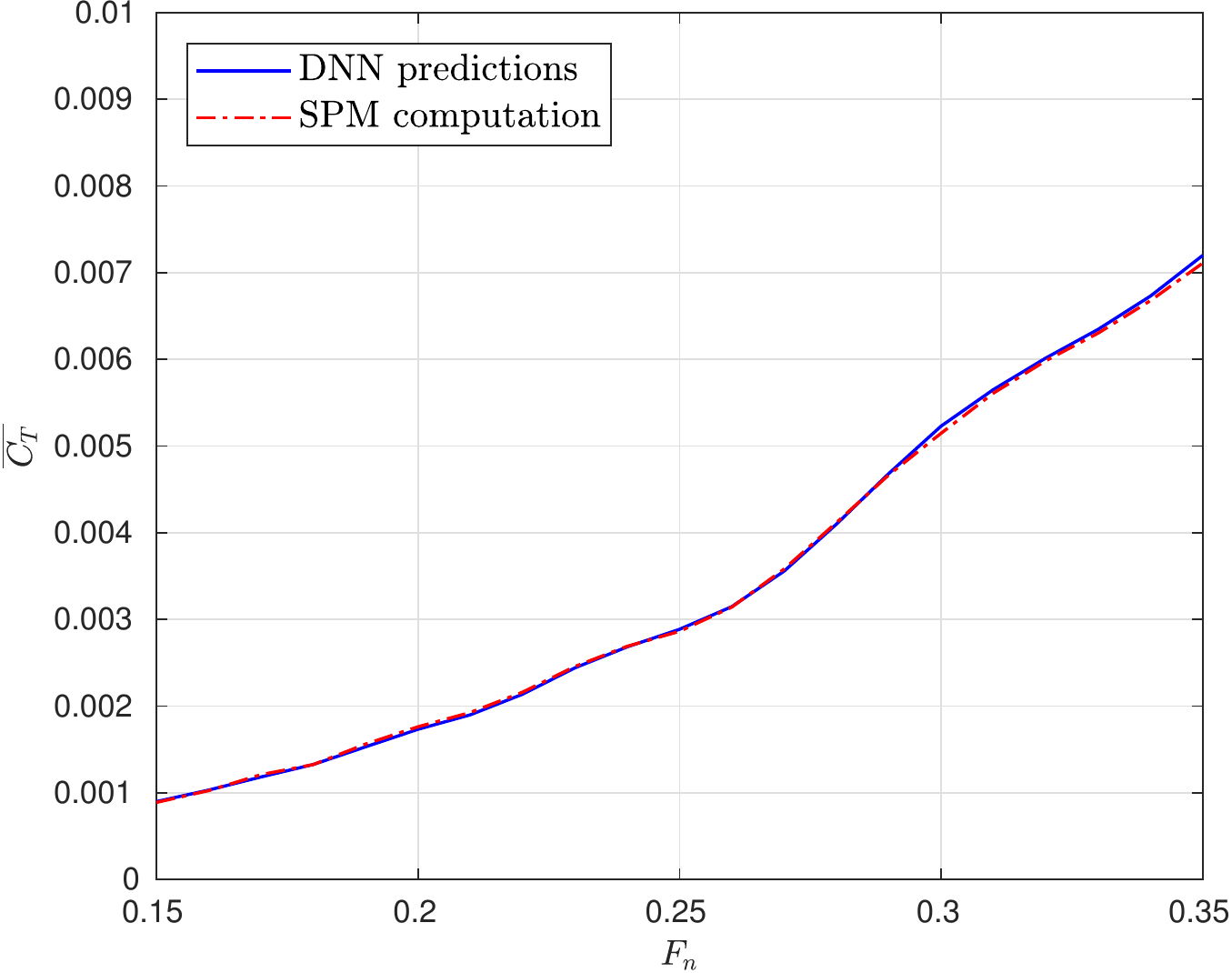}
        \caption{}
        \label{fig:4.2b}
    \end{subfigure}}
    {\caption{Comparisons between the results of DNN and SPM computation for selected hull forms. (a) $\overline{\bm{\lambda}} = [0.073, 0.758, 0.050]$, $L/B = 8.249$, $B/T = 2.751$. (b) $\overline{\bm{\lambda}} = [0.766, 0.174, 0.503]$, $L/B = 7.704$, $B/T = 3.466$.}\label{fig:4.2}}
\end{figure}

The DNN predictions of $\overline{C}_T$ for selected hull forms, which are not included in the training samples, are plotted against the corresponding SPM computational results in Fig.~\ref{fig:4.2}. Very good agreement is observed between DNN predictions and the computational results; therefore, the DNN is adequately trained to accurately predict the merit coefficient of a vessel based on its hull-form parameters, Froude number, and Reynolds number.

\subsection{Evaluation of Vessel Performance and the Search for Optimal Hull Form using DNN}
Since a large amount of hull forms can be quickly and simultaneously evaluated by the DNN forward propagation, a straightforward Monte-Carlo search for optimal hull form with the lowest value of $\beta$ subject to prescribed operational requirements is carried out, and the procedures are described as follows:

\begin{enumerate}
	\item Prescribe the length of the target hull form and the lower and upper search limits for each of the hull-form parameters. Specify the Froude number range of interest, \ie~$F_{n_l}$ and $F_{n_u}$, and the number of equally spaced Froude numbers $n_{U}$ to be used in this range. Based on the length of the hull and the Froude number, the Reynolds number can be determined accordingly.
    \item Generate $n_1$ hull forms within the design space, by randomly choosing a value for each of the five hull-form parameters within the limits specified in the previous step.
    \item Submit the generated hull-form parameters, Froude numbers, and Reynolds numbers to the DNN and compute the corresponding values of $\overline{C}_T$  by forward propagation.
    \item Based on the Froude number and resulting values of $\overline{C}_T$, compute the value of $\beta$ with prescribed weight function $w(F_n)$ for each generated hull form by Eq.~\eqref{eq:3.6}.
    \item Find the minimum value $\beta_{min}$ and the associated optimal hull form.
    \item Repeat Step 2 to 5 for $k$ times until the value of $\beta_{min}$ exhibits insignificant changes. In the meantime, the search range for each hull-form parameters can be shrank down based on the intermediate optimal solution from the last iteration, so that the optimal hull form can be more accurately determined.
\end{enumerate}

The above search process is straightforward and highly customizable, since any restrictions on the hull form and forward speed can be easily incorporated by specifying searching limits for each of the inputs to the DNN. To ensure the accuracy of the optimization search, large values of $n_U$ and $n_1$ can be adopted to provide more data points, and the entire search effort can be repeated multiple times to ensure the convergence of the result.  

In addition, to refine the results of the search and ensure that global optimum is indeed achieved, an additional search can be conducted targeting each individual hull-form parameter, and the relevant procedures are provided as follows:

\begin{enumerate}
	\item Pick one of the five hull-form parameters to vary within a search range of interest, while the rest of the parameters are kept constant as those of the last iteration. As a result, $n_2$ new hull forms are generated in the process.
	\item Evaluate $\beta$ for the newly-generated hull forms. Update the value of $\beta_{min}$ and the associated optimal hull form if a better one is discovered.
	\item Select another hull-form parameter and repeat Steps 1 and 2, until the search is completed for all hull-form parameters.
\end{enumerate} 

The refined search described above can also be repeated several times for convergence. In addition, the sequence of parameters being optimized can also be randomized between each iteration.  

\section{Results and Discussions}
For demonstration purposes, two optimization searches are conducted targeting two different vessel designs with distinct length and operating speed ranges, they are:

\begin{enumerate}
	\item A 300-meter long vessel operating with Froude numbers between $[0.20, 0.24]$ (21.1 - 25.3 knots). The dimension and the target speed range is approximately consistent with a typical ocean-going container ship with a capacity 300 to 5000 Twenty-Foot Equivalent Units (TEUs).
	\item A 170-meter long vessel operating with Froude numbers between $[0.26, 0.30]$ (20.6 - 23.8 knots). The dimension and operating speed are selected based on those of feeder ships, which are container ships with a smaller shipping capacity between 300 to 3000 TEUs for collecting shipping containers from different ports and delivering them to central container terminals for further transportation~\cite{ng2008}. 
\end{enumerate} 

\noindent The design space is outlined in Table~\ref{table:5.1} for the search. The weight function $w(F_n)$ used in the computation of $\beta$ by Eq.~\eqref{eq:3.6} is taken as 

\begin{equation}
\label{eq:5.1}
w = \frac{1}{F_{n_u} - F_{n_l}},
\end{equation}

\noindent \ie~the distribution of operating speed is uniform in the investigated ranges.

\begin{table}[!htbp]
	\centering
	\caption{Ranges for hull-form parameters in the optimization search.}
	\begin{tabular}{|c|c|c|c|c|c|}
		\hline 
		Hull-form parameters & $\overline{\lambda}_1$ & $\overline{\lambda}_2$ & $\overline{\lambda}_3$ & $L/B$ & $B/T$ \\ 
		\hline 
		Minimum & 0 & 0 & 0 & 7 & 2.0\\ 
		\hline 
		Maximum & 1 & 1 & 1 & 9 & 3.1 \\ 
		\hline 
	\end{tabular}  
	\label{table:5.1}
\end{table}

\subsection{Optimal Hull Form for $F_n\in [0.20, 0.24]$ and $L = 300$m}
\begin{figure}[!htbp]
\centerline{
	\begin{subfigure}{.55\textwidth}
			\includegraphics[width=1\columnwidth]{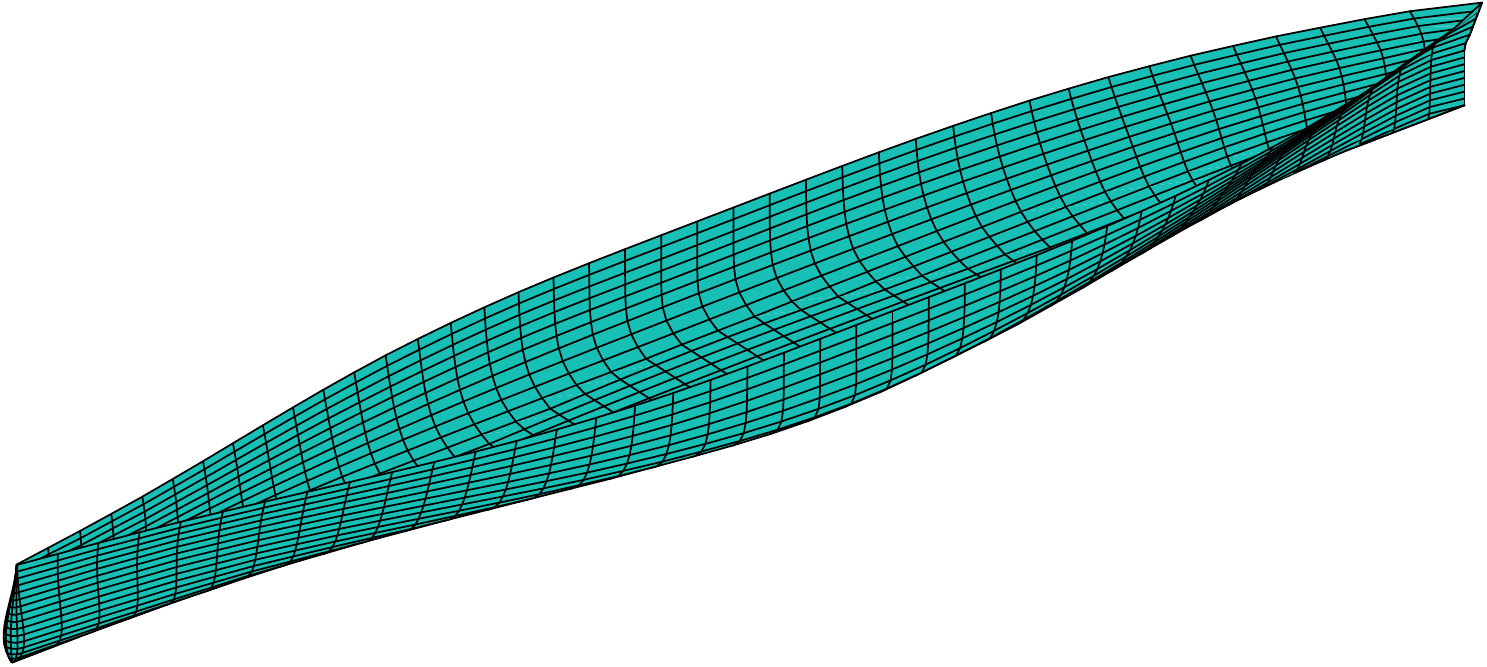}
			\caption{Three-dimensional view.}
			\label{fig:5.1a}
		\end{subfigure}\;\;\;
		\begin{subfigure}{0.45\textwidth}
			\includegraphics[width=1\columnwidth]{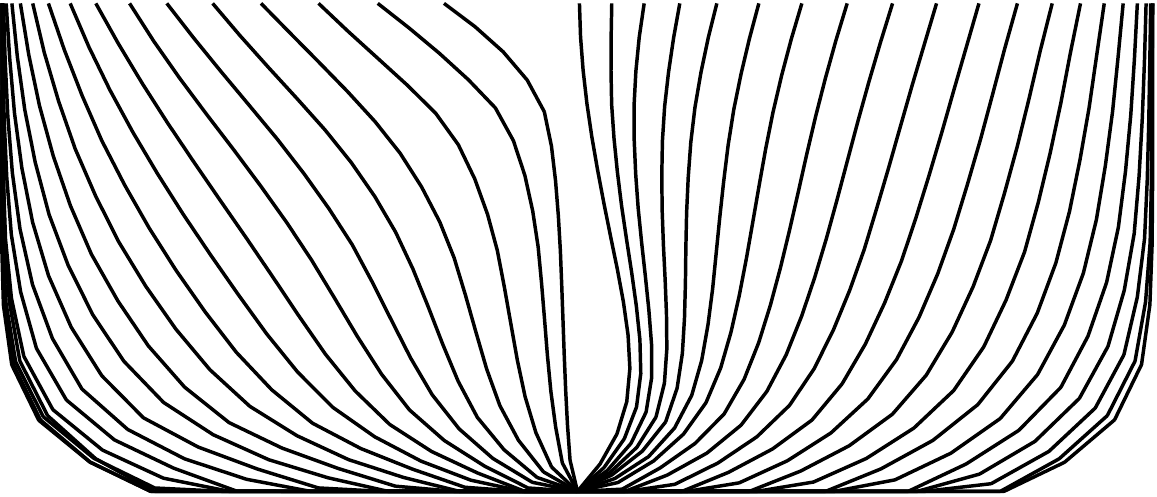}
			\caption{Body plan.}
			\label{fig:5.1b}
		\end{subfigure}}
		{\caption{Optimal hull form for $F_n\in [0.20, 0.24]$ and $L = 300$m.}\label{fig:5.1}}
\end{figure}

With the GPU-accelerated forward propagation on the DNN, the search for optimal hull form with $n_1 = n_2 = 3000$, $k = 15$, and $n_U = 5$ takes less than 30 seconds on an Nvidia GTX 1080 GPU. The optimal hull form is shown in Fig.~\ref{fig:5.1}, and the associated hull-form parameters and geometric coefficients are given in Table~\ref{table:5.2}. The resulting optimal hull form shares great similarity with the S175 container ship. Within the evaluated speed range, the values of $\overline{C}_T$ and $\beta$ of the optimal hull form is also similar to those of the S175 container ship with slight improvement, as shown in Fig.~\ref{fig:5.5a} and Table~\ref{table:5.3}. The resulting geometry of the optimal hull form is overall consistent with the designs of modern ocean-going container ships.

\begin{table}[!t]
	\centering
	\caption{Hull-form parameters for the optimal hull form of $F_n\in [0.20, 0.24]$ and $L = 300$m.}
	\begin{tabular}{|c|c|c|c|c|c|c|c|}
		\hline 
		 $\overline{\lambda}_1$ & $\overline{\lambda}_2$ & $\overline{\lambda}_3$ & $L/B$ & $B/T$ & $C_B$ & $C_P$ & $L/\Vol^{1/3}$\\ 
		\hline 
		 0.6818 & 0.0001 & 0.1339 & 6.99 & 2.36 & 0.570 & 0.587 & 5.874\\ 
		\hline 
	\end{tabular}  	
	\label{table:5.2}
\end{table}

\begin{table}[!htbp]
	\centering
	\caption{Values of $\beta$ for different hull forms for $F_n\in [0.20,0.24]$ (lower is better).}
	\begin{tabular}{|c|c|c|}
	\hline 
	\multirow{2}{*}{Hull forms} & \multirow{2}{*}{$\beta$ }& Difference with respect to \\ 
	& & optimal hull form in $\%$ \\ 
	\hline 
	Series 60 $C_B = 0.6$ & 0.0017 & 12.68 \\ 
	\hline 
	Series 60 $C_B = 0.7$ & 0.0020 & 38.36 \\ 
	\hline 
	S175 container ship & 0.0015 & 1.87 \\ 
	\hline 
	Wigley hull & 0.0021 & 43.42 \\ 
	\hline 
	optimal hull form & 0.0015 & 0 \\ 
	\hline 
	\end{tabular} 
	\label{table:5.3}
\end{table}

\subsection{Optimal Hull Form for $F_n\in [0.26, 0.3]$ and $L = 170$m}
\begin{figure}[!htbp]
\centerline{
	\begin{subfigure}{.55\textwidth}
			\includegraphics[width=1\columnwidth]{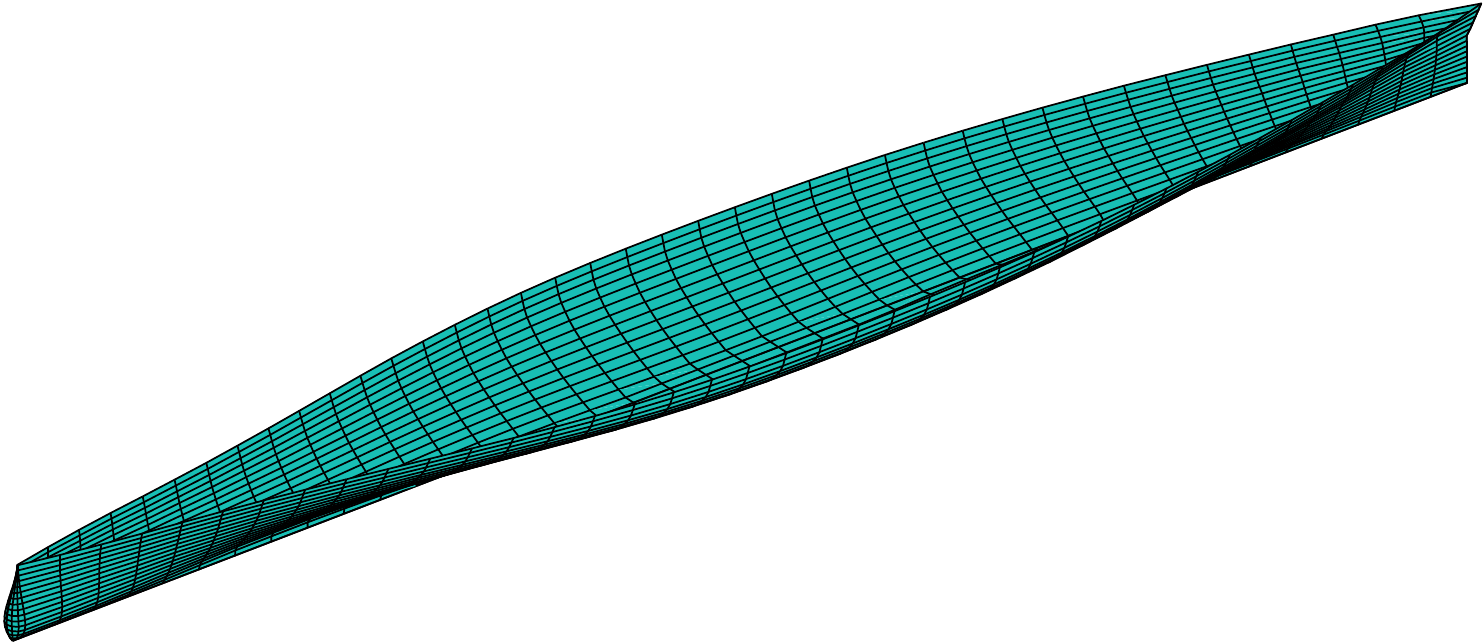}
			\caption{Three-dimensional view.}
			\label{fig:5.2a}
		\end{subfigure}\;\;\;
		\begin{subfigure}{0.45\textwidth}
			\includegraphics[width=1\columnwidth]{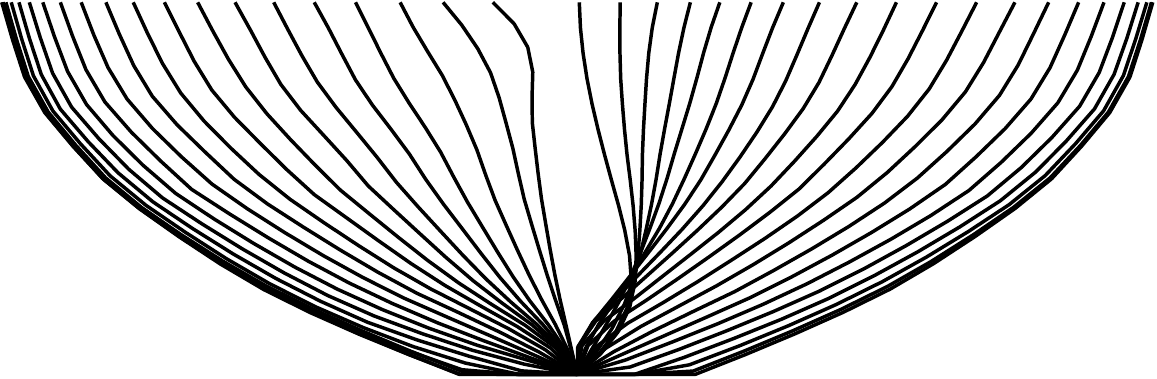}
			\caption{Body plan.}
			\label{fig:5.2b}
		\end{subfigure}}
		{\caption{Optimal hull form for $F_n\in [0.26, 0.30]$ and $L = 170$m.}\label{fig:5.2}}
\end{figure}

\begin{figure}[!t]
\centerline{
	\begin{subfigure}{0.5\textwidth}
			\includegraphics[width=1\columnwidth]{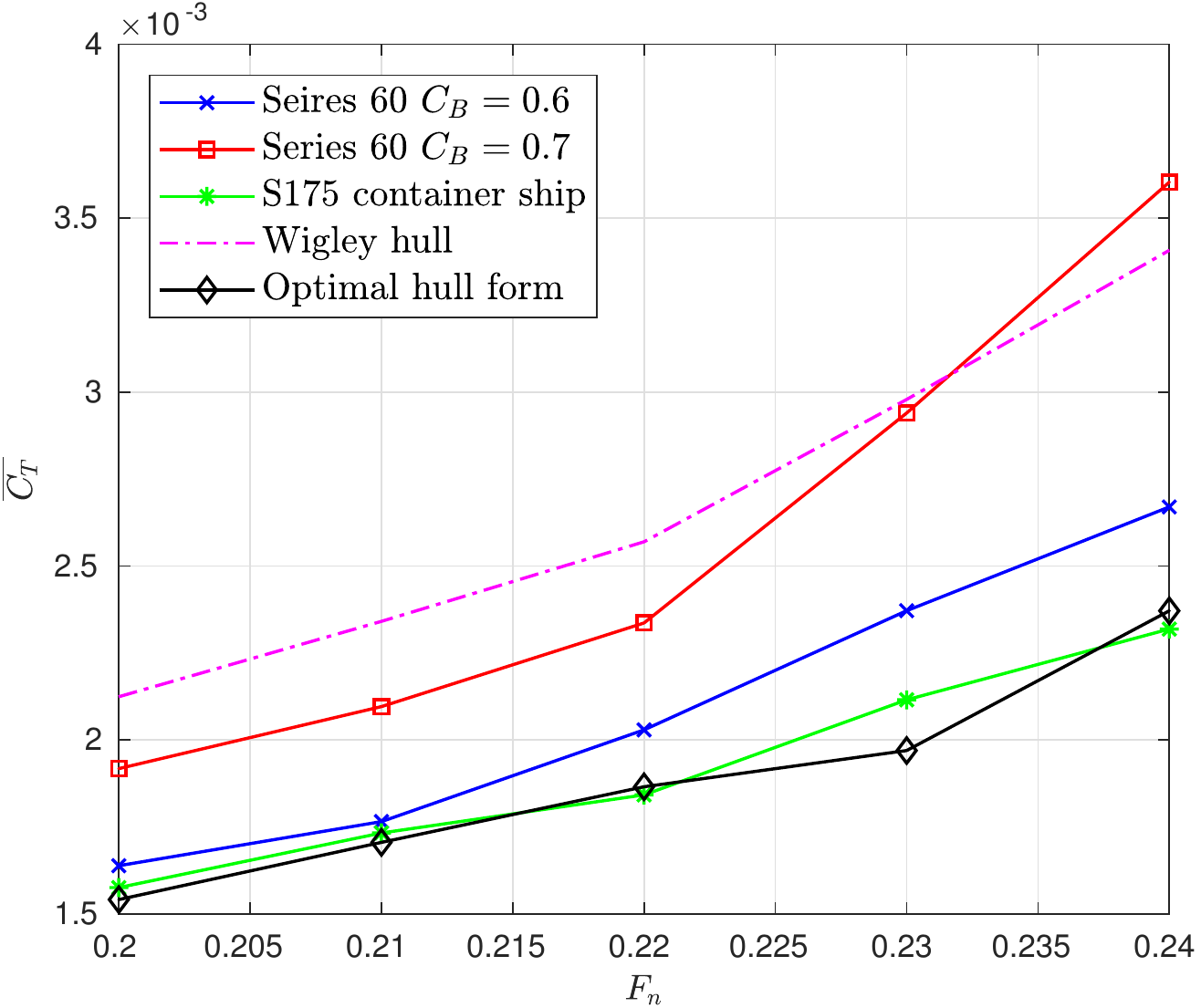} 
			\caption{$F_n\in [0.20, 0.24]$ with $L = 300$m}
			\label{fig:5.5a}
		\end{subfigure}%
		\begin{subfigure}{0.5\textwidth}
			\includegraphics[width=1\columnwidth]{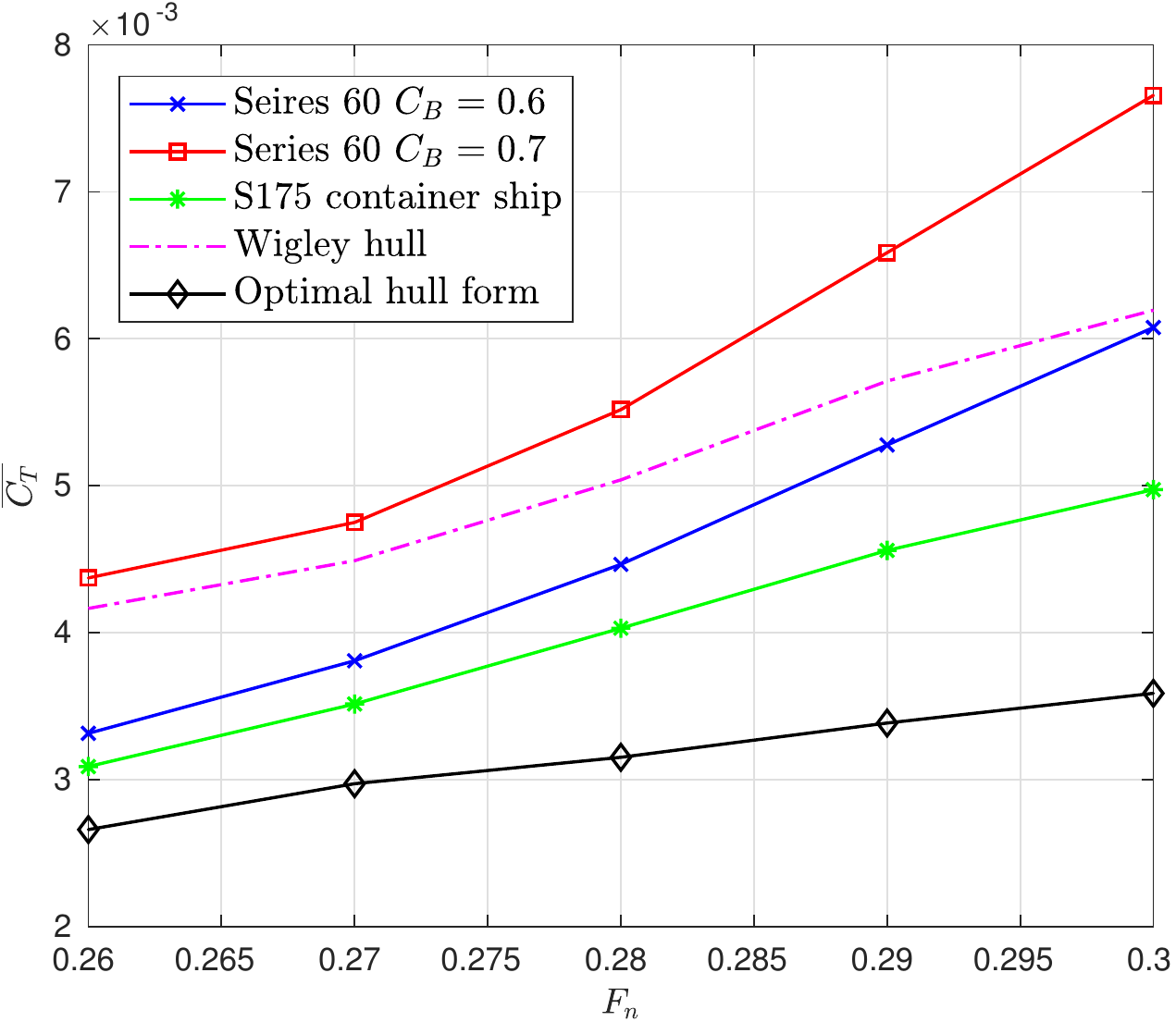} 
			\caption{$F_n\in [0.26, 0.30]$ with $L = 170$m}
			\label{fig:5.5b}
		\end{subfigure}}
		{\caption{Comparisons of the values of $\overline{C}_T$  for different hull forms of length (a) $L=300$m and (b) $L=170$m.}\label{fig:5.5}}
\end{figure}

The optimal hull form for $F_n\in [0.26, 0.3]$ and $L = 170$m is shown in Fig.~\ref{fig:5.2}, and the associated hull-form parameters and geometric coefficients are given in Table~\ref{table:5.4}. The acquired optimal hull form resembles a combination of S175 container ship and the Wigley hull with a round midship section. In addition, the discovered small length-to-beam ratio and large beam-to-draft ratio are consistent with the general design philosophy of feeder vessels. Furthermore, it is shown in Fig.~\ref{fig:5.5b} and Table~\ref{table:5.5} that, within the evaluated speed range, the acquired optimal hull form considerably outperforms all four parent hulls with smaller values of $\overline{C}_T$  and $\beta$.

\begin{table}[!htbp]
	\centering
	\caption{Hull-form parameters for the optimal hull form of $F_n\in [0.26, 0.3]$ and $L = 170$m.}
	\begin{tabular}{|c|c|c|c|c|c|c|c|}
		\hline 
		 $\overline{\lambda}_1$ & $\overline{\lambda}_2$ & $\overline{\lambda}_3$ & $L/B$ & $B/T$ & $C_B$ & $C_P$ & $L/\Vol^{1/3}$\\ 
		\hline 
		 0.0019 & 0.0001 & 0.0009 & 6.90 & 3.10 & 0.411 & 0.559 & 7.18\\ 
		\hline 
	\end{tabular}  
	\label{table:5.4}
\end{table}

\begin{table}[!htbp]
	\centering
	\caption{Values of $\beta$ for different hull forms for $F_n\in [0.26,0.3]$ (lower is better).}
	\begin{tabular}{|c|c|c|}
	\hline 
	\multirow{2}{*}{Hull forms} & \multirow{2}{*}{$\beta$ }& Difference with respect to \\ 
	& & optimal hull form in $\%$ \\ 
	\hline 
	Series 60 $C_B = 0.6$ & 0.0046 & 44.44\\ 
	\hline 
	Series 60 $C_B = 0.7$ & 0.0057 & 81.04 \\ 
	\hline 
	S175 container ship & 0.0040 & 27.74 \\ 
	\hline 
	Wigley hull & 0.0051 & 61.67 \\ 
	\hline 
	optimal hull form & 0.0032 & 0 \\ 
	\hline 
	\end{tabular} 
	\label{table:5.5}
\end{table}

\section{Conclusions and Future Work}
With the help of PCA, the complex geometries of a group of different ship hulls can be easily compressed into and recovered from several principal scores. Compared to conventional geometric coefficients, the principal scores systematically identified by PCA enables more accurate representation of the hull geometry. Furthermore, by changing the values of the principal scores, new hull forms that inherits geometric features of the parent hulls can be easily generated without directly manipulating the large number of offset values.

The GPU-accelerated SPM computation is used to efficiently compute the calm-water resistances of a large number of hull geometries generated through manipulation of the principal scores. The resulting dataset is used to train a DNN, with which large quantities of different hull forms can be evaluated at negligible computing costs with a similar level of accuracy as the SPM computation itself. Therefore, the DNN-based search for optimal hull forms can be performed with great speed and flexibility. The resulting optimal hull form combines the geometric characteristics of the parent hulls to better suit the specified design requirements.

In short, the present study marks a successful and novel application of machine learning in the field of ship design and optimization. It should be noticed that the potential of the present approach can be further explored in several directions. First of all, as mentioned earlier, more existing vessels can be included as parent hulls in the PCA to generate hull forms with more variations, thus expanding the design space. Secondly, results of high-fidelity hydrodynamic analysis, such as those of viscous-fluid CFD simulations and model-scale experiments can be included in the training and test samples to further improve the reliability of the DNN predictions. Last but not least, other important aspects of ship performances such as seakeeping and maneuverability can be evaluated in a similar manner and incorporated into the optimization criteria, so that a more comprehensive evaluation and optimization of hull geometry can be achieved.

\bibliographystyle{ieeetr}
\bibliography{pcabib}
\end{document}